\documentclass{article}
\PassOptionsToPackage{accepted}{mlsys2025}
\usepackage{microtype}
\usepackage{graphicx}
\usepackage{subfigure}
\usepackage{booktabs}
\usepackage{amsmath}
\usepackage{xspace}
\usepackage{bm}
\usepackage{multirow}
\usepackage{threeparttable}
\usepackage{mathrsfs}
\usepackage{pifont}
\usepackage{dblfloatfix}
\usepackage[above]{placeins}
\usepackage{listings}
\usepackage{xcolor}
\usepackage{tabularx}
\usepackage[normalem]{ulem}
\usepackage{xr-hyper}
\usepackage{float}
\usepackage{xurl}
\usepackage{hyperref}

\usepackage{mlsys2025}
\hypersetup{pdfsubject={MLSys 2027 submission}}

\AtBeginDocument{%
  \providecommand\BibTeX{{%
    \normalfont B\kern-0.5em{\scshape i\kern-0.25em b}\kern-0.8em\TeX}}}
\newcommand{\name}{CrossInfer\xspace}

\ifodd 1

\else

\fi

\makeatletter
\renewcommand{\Notice@String}{}
\makeatother
\hypersetup{pdfsubject={Preprint}}
\mlsystitlerunning{Shared Context, Separate Engines}

\begin{document}

\twocolumn[
\mlsystitle{Shared Context, Separate Engines: Stateful Streaming Inference for Multimodal LLMs on Edge SoCs}

\begin{mlsysauthorlist}
\mlsysauthor{Yilong Li}{uw}
\mlsysauthor{Chengpo Yan}{uw}
\mlsysauthor{Yijing Zeng}{meta}
\mlsysauthor{Hanlei Wu}{cmu}
\mlsysauthor{Mingjun Song}{uw}
\mlsysauthor{Suman Banerjee}{uw}
\end{mlsysauthorlist}
\mlsysaffiliation{uw}{University of Wisconsin--Madison}
\mlsysaffiliation{meta}{Meta}
\mlsysaffiliation{cmu}{Carnegie Mellon University}

\begin{center}
\small Code: \url{https://github.com/jimmy-yilong-li/CrossInfer}
\end{center}

\vskip 0.3in
\begin{abstract}
Streaming multimodal large language models (MLLMs) must take in new
video frames while answering questions about what they have already
seen. On an edge SoC, if one engine does both, ingestion and answering
compete for its time while the other engines sit idle. To take over a
query, another engine must first import or rebuild the accumulated
history, because each runtime keeps it as KV state in its own format.
\name lets one engine answer from history that another engine built.
Each query reads the history as it stood when the query was asked, so
one engine can keep ingesting while another answers. The CPU and GPU
can also decode answers to different queries simultaneously.
Phase-level state continuation lets an engine prefill a query, pass its
KV state to another engine for decoding, and return to ingestion.
State-Driven Execution Planning (SDEP) chooses where each query is
prefilled and decoded, based on the state each engine holds and the work
queued on it.
Across M4 Max, RK3588, and QCS6490, \name outperforms running on a
single engine, as llama.cpp and MLX do. On a live M4 Max stream, SDEP cuts p95 time to first token
by $76\%$, and every query receives its first token within one second.
Under backlog, it raises request throughput by up to $35\%$. For two
concurrent video streams on RK3588, SDEP delivers $19.6\%$ higher
request throughput than a single engine and $27.2\%$ more than
running both engines in parallel without shared state.
\name delivers up to $2.48\times$ the throughput of the same engines
run sequentially. Every answer matches sequential execution token for token.

\end{abstract}
]

\begingroup
\let\thefootnote\relax
\begin{NoHyper}
\footnotetext{\hspace*{-\footnotesep}%
\textsuperscript{1}University of Wisconsin--Madison.
\textsuperscript{2}Meta.
\textsuperscript{3}Carnegie Mellon University.}
\end{NoHyper}
\endgroup

\section{Introduction}
\label{sec:intro}
Multimodal large language models (MLLMs) integrate images, audio, and text, allowing edge devices to answer questions about a video as it unfolds~\cite{liu2023llava,liu2024llavanext,llava_onevision,marafioti2025smolvlm,Qwen2VL,Qwen2.5-VL,Qwen3-VL,qwen35}. Streaming asks the device to do two things at once: add each new frame to the model's history, and answer each query from that history without delay. An edge SoC has a CPU, a GPU, and an NPU for this work, yet when one of them does both, each task stalls the other while the rest of the chip sits idle. Another engine cannot simply step in, because the history lives as KV state inside the runtime of the engine that built it.

\begin{figure*}[thb]
    \centering
    \includegraphics[width=\textwidth]{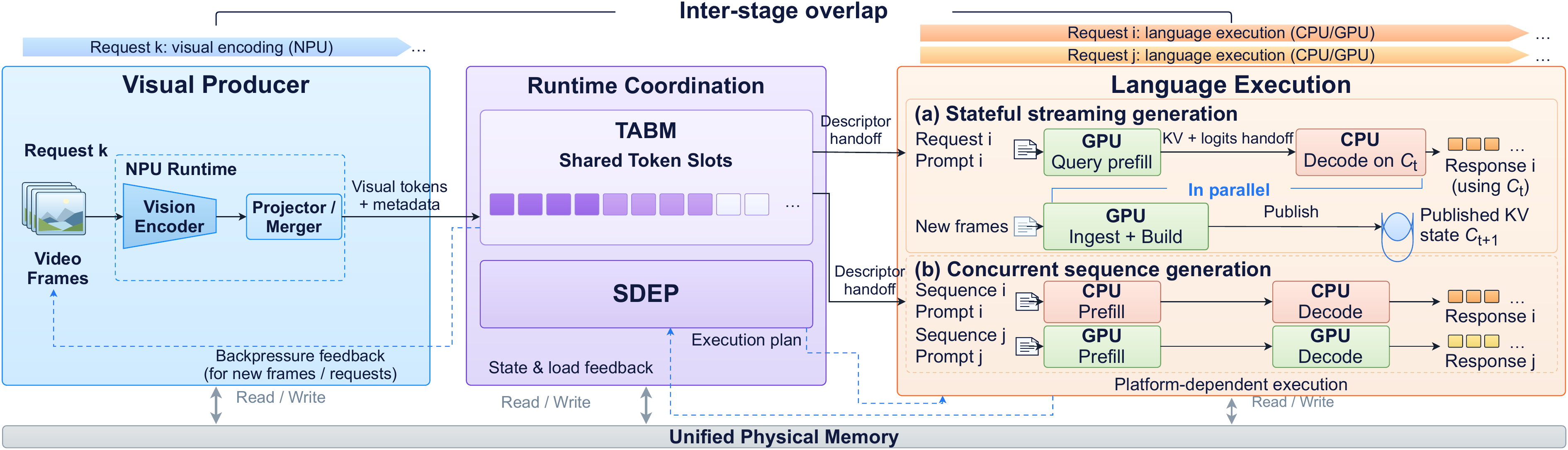}
    \caption{Overview of \name. (a) \textbf{Stateful Streaming Generation:} One engine ingests new observations while another generates responses from published context, using state handoff. (b) \textbf{Concurrent Sequence Generation:} CPU and GPU execute ready requests in parallel, with SDEP selecting their execution paths.}
    \label{fig:intro}
\end{figure*}

Recent systems improve multimodal inference through heterogeneous execution and context reuse. NanoMind~\cite{nanomind} dynamically places model modules across processors and exchanges visual tokens between them. TriInfer~\cite{triinfer} separates encoding, prefill, and decoding and reallocates resources across these stages. For streaming inputs, ReKV~\cite{rekv} reuses video KV caches to separate video analysis from question answering, while Stream2LLM~\cite{stream2llm} overlaps context retrieval with prefill and adapts scheduling to arriving context. These systems leave open how to continue a stream's language state across heterogeneous engines on one SoC while preserving each query's history and sustaining ingestion.

Running the CPU and GPU together can improve throughput, but contention for shared memory bandwidth can also make inference slower. Batch-one decoding is fastest on the M4 Max GPU but on the RK3588 CPU (Figure~\ref{fig:engine_cost}). On M4 Max, adding a second language engine raises throughput by up to $29\%$ or reduces it by $45\%$, depending on the workload. Longer prefill reduces the gain there but increases it on RK3588 (Figure~\ref{fig:lane_boundary}). Context reuse~\cite{promptcache,sglang} avoids recomputing history but retains it as backend-specific KV state and positional metadata. Importing or reconstructing this state on an idle engine can cost more than waiting for a busy engine that already holds it. Assigning an entire query to one engine also requires it to perform query prefill, even if its strength is decoding.

We present \name, a system for stateful MLLM inference across edge accelerators (Figure~\ref{fig:intro}). It separates logical history from engine-specific KV state so queries can change engines while keeping their histories fixed. \emph{Stateful Streaming Generation} retains and reuses video context across queries. The CPU and GPU share the work: one processes new video frames while the other generates answers from the accumulated history. \emph{Concurrent Sequence Generation} also lets the two engines answer different queries simultaneously, within or across video streams. They can divide the work within a query as well. Through phase-level state continuation, the GPU prefills the query and passes its KV state to the CPU for decoding, then resumes ingestion. Meanwhile, the NPU encodes incoming frames for the next context update. State-Driven Execution Planning (SDEP) selects the engines for ingestion, query prefill, and decoding, and determines how many queries each engine handles.

\begin{figure}[t]
\centering
\subfigure[M4 Max: p95 TTFT]{%
\includegraphics{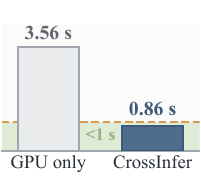}%
\label{fig:intro_ttft}%
}\hfill
\subfigure[Throughput Speedup~(x1)]{%
\includegraphics{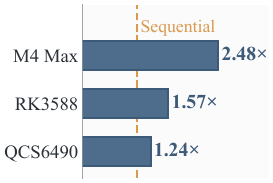}%
\label{fig:intro_throughput}%
}
\caption{(a) On a live M4 Max stream, \name cuts p95 time to first
token by $76\%$ relative to GPU-only execution as in
llama.cpp~\cite{llama_cpp} and MLX~\cite{mlx2023}. Every first
token arrives within one second. (b) Stateful Streaming Generation
reaches up to $2.48\times$ the throughput of the same engines run
sequentially.}
\label{fig:intro_speedup}
\end{figure}

We implement \name on M4 Max, RK3588, and QCS6490 (Section~\ref{sec:impl}). On a live M4 Max stream, SDEP cuts p95 time to first token (TTFT) by $76\%$ relative to GPU-only execution, and every first token arrives within one second. Under backlog, throughput exceeds the same baseline by up to $35\%$ (Section~\ref{ssec:eval:e2e}). For the 27B Qwen3.6 and Qwen3.8 models~\cite{qwen36,qwen38}, SDEP raises throughput by up to $17.5\%$ over GPU-only execution. Cross-engine state sharing adds gains beyond context reuse. On two RK3588 streams, SDEP delivers $19.6\%$ higher request throughput and $34.9\%$ lower mean TTFT than a single engine, and $27.2\%$ higher throughput than two engines without shared state. Importing KV on RK3588 reduces first-token latency by a factor of $4.6$ compared with recomputing the prefix. Across nine RK3588 history/backlog settings, SDEP selects the faster engine each time (Figure~\ref{fig:context_placement}).

Stateful Streaming Generation yields $2.48\times$ the throughput of the same engines run sequentially on M4 Max and $56.7\%$ more on RK3588 (Figure~\ref{fig:intro_speedup}). Concurrent Sequence Generation improves throughput over GPU-only on RK3588 by up to $52.0\%$ in a concurrency sweep (Section~\ref{ssec:eval:concurrent}). Both modes beat the best single engine by up to $37.2\%$ for OneVision and FastVLM~\cite{llava_onevision,fastvlm2025} on RK3588 and on QCS6490, which also encodes every frame on its NPU. Vision--language overlap adds up to $72.4\%$ throughput across four Qwen3.5 models from 0.8B to 9B, relative to the same pipeline with overlap disabled. On M4 Max, \name delivers $29.6\%$ higher throughput than GPU-only MLX~\cite{mlx2023} on a 64-frame video (Table~\ref{tab:video_mme}) and uses $23.6\%$ less energy than llama.cpp~\cite{llama_cpp} for a single Qwen2.5-VL-3B request.

This paper makes three contributions:
\begin{itemize}
    \item \textbf{Cross-accelerator execution of stateful MLLM inference.}
    We unify stage overlap and request concurrency so engines can update
    a stream's history while answering queries and serve ready requests
    in parallel within or across streams.

    \item \textbf{Versioned state continuation.}
    Versioned context and explicit token and state interfaces let
    any language engine continue a stream's history. Context handoff and
    phase-level state continuation preserve each query's history binding
    and the computation completed before handoff.

    \item \textbf{State-Driven Execution Planning.}
    SDEP chooses the ingestion engine using state location and execution
    costs, and selects query paths by predicted completion time.
    It adjusts how many queries each engine answers according to
    arrival rates and the time that handoff frees for ingestion.
    We show how stage costs and co-running interference, which differ across SoCs,
    determine which engines to combine.
\end{itemize}

\section{Background and Motivation}
\label{sec:motivation}
\subsection{MLLM Inference on the Edge}
\label{ssec:mot:edge}

\begin{figure}[htb]
\centering
\includegraphics[width=\columnwidth]{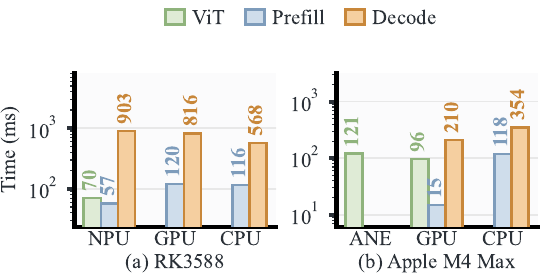}\\[3pt]
\includegraphics[width=\columnwidth]{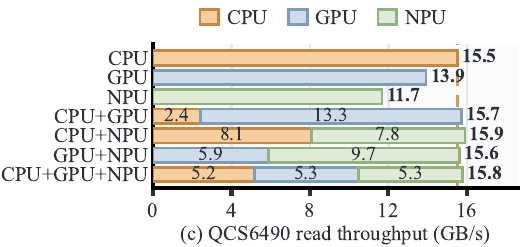}
\caption{Stage latency for one image (ViT~\cite{vit}) and 32-token
prefill and decode (log scale). (a) RK3588: CLIP ViT-B/32 and
Qwen2.5-0.5B-Instruct~\cite{qwen25}. (b) M4 Max: Qwen2.5-VL-3B
(Q4\_K\_M), with Core~ML~\cite{apple_coreml} vision on the ANE.
(c) QCS6490 effective read throughput for 256\,MiB inputs; the dashed
line marks CPU-only throughput (Appendix~\ref{app:qcs6490_bandwidth}).}
\label{fig:engine_cost}
\end{figure}

Edge MLLMs often run on a single accelerator although modern SoCs offer
several. GPU-only MLX, for example, keeps the M4 Max GPU $85$--$99\%$
busy while the Apple Neural Engine (ANE)~\cite{apple_ane_report} sits
idle (Table~\ref{tab:video_mme}). Yet MLLM stages favor different
engines: the fastest decode engine differs by SoC
(Figure~\ref{fig:engine_cost}), and on RK3588 generating 32 tokens takes
$568$--$903$\,ms across engines against $70$\,ms for CLIP~\cite{CLIP}
ViT-B/32 encoding on the NPU. Prior work separates LLM prefill and
decoding~\cite{splitwise,distserve} or exploits heterogeneous
processors~\cite{llmnpu,powerinfer2,heteroinfer}, and
NanoMind~\cite{nanomind} pairs a custom board with a commodity SoC for
cross-accelerator MLLM inference. Shared memory~\cite{apple_m1_2020} does
not make one engine's state usable by another: each runtime keeps its
own buffers and KV-cache layout.

Engines also interfere when they run together: co-running slows RK3588 CPU and Mali
GPU~\cite{arm_mali_g610} decoding by $26\%$ and $11\%$
(Figure~\ref{fig:lane_interference}). On QCS6490, co-running engines
share about $16$\,GB/s of read bandwidth unevenly: with CPU and GPU
active, the GPU keeps $13.3$\,GB/s while the CPU falls from $15.5$ to
$2.4$\,GB/s (Figure~\ref{fig:engine_cost}(c)). On Snapdragon 8
Gen 3~\cite{qualcomm_snapdragon8gen3}, HeteroInfer~\cite{heteroinfer}
reports that running the GPU and NPU together raises aggregate decoding
bandwidth from $40$--$45$ to about $60$\,GB/s, so bandwidth sharing is
itself platform-specific.

\subsection{Streaming Applications}
\label{ssec:mot:streaming}
Streaming MLLMs expose parallelism that single-engine execution leaves unused. In online video understanding~\cite{ovo_bench}, new frames continuously extend the video context while queries refer to the history accumulated so far. On one engine, ingestion and generation compete for the same execution capacity, so each delays the other. Yet later frames could be encoded and ingested while an earlier query is answered. Ready queries can also run concurrently, whether they share a visual history or belong to separate streams.
Exploiting this parallelism requires managing context that evolves throughout the session. Each answer must read a consistent version of the history, even as new frames arrive, while concurrent answers maintain independent generation state. Pausing ingestion sacrifices overlap, whereas copying the full history for each answer wastes memory. Execution on another engine requires importing compatible KV state or reconstructing it through prefill, so placement depends on context availability as well as engine load~\cite{preble}. Requiring an engine to prefill its own queries delays decoding when another engine could prepare the same state faster.

\section{Design}
\label{sec:design}
The engines of an edge SoC can cooperate on MLLM inference at three
levels: assigning stages to suitable engines, running engines
concurrently, and coordinating a stream's state across engines.
\name lets a query keep its bound history while its execution moves
across engines. Versions and private answer branches separate stream
updates from answering. State handoff preserves completed
computation when another engine takes over after context ingestion or
query prefill.
SDEP uses engine costs and state availability to choose where each phase
runs and distribute arriving queries among lanes.

\subsection{Execution Model}
\label{sec:execution-model}

\begin{figure*}[t]
\centering
\subfigure[Stateful Streaming Generation]{%
\includegraphics[width=0.42\textwidth]{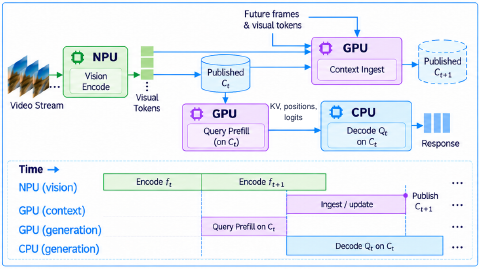}%
\label{fig:exec_streaming}%
}\qquad
\subfigure[Concurrent Sequence Generation]{%
\includegraphics[width=0.42\textwidth]{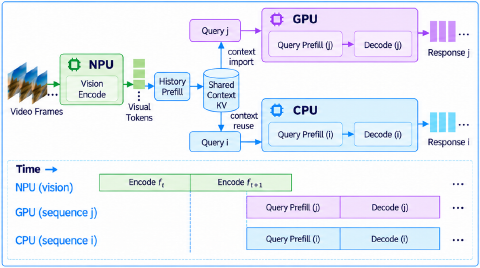}%
\label{fig:exec_concurrent}%
}
\caption{Execution modes in \name. (a) After GPU query prefill, the CPU
decodes $Q_t$ while the GPU builds $C_{t+1}$. (b) Two lanes answer
different queries over $C_t$: the CPU reuses it in place and the GPU
imports it.}
\label{fig:exec_models}
\end{figure*}

A stream supplies observations in temporal order, each an image or a
group of video frames. Visual encoding produces embeddings for each
observation. Context ingestion adds them to the stream history, query
prefill processes a question on a private answer branch, and decoding
extends that branch with answer tokens. Each completed history is
published as a context version: we write $f_t$ for the
$t$-th observation, $C_t$ for the version that ends with it, and $Q_t$
for a query bound to $C_t$. A \emph{language lane} is a runtime worker
assigned to one engine. It keeps its own KV state and executes the
prefill and decoding work assigned to it. A lane stays on one engine,
but a query can continue on another lane at a supported handoff boundary. \name supports two modes
(Figure~\ref{fig:exec_models}).

\noindent\textbf{Stateful Streaming Generation.}
One lane ingests new observations and publishes each context version
while another lane answers from a published version. In
Figure~\ref{fig:exec_streaming}, the GPU prefills $Q_t$ on a private
branch of $C_t$ and hands the branch to the CPU, which decodes while the
GPU builds $C_{t+1}$.

\noindent\textbf{Concurrent Sequence Generation.}
CPU and GPU lanes execute ready requests concurrently. Requests bound to the same published context reuse its history while maintaining private answer branches. In Figure~\ref{fig:exec_concurrent}, the CPU reuses $C_t$ locally while the GPU imports it, and each answers its own query. Requests within a stream may bind different versions, while separate streams maintain independent histories. Each lane can batch requests, while \name chooses the engines and state location for each request.
Both modes can operate in one serving session, and phase-level state
continuation can split a query's prefill and decoding across lanes in
either mode
(Section~\ref{sec:token-state-interfaces}).

\noindent\textbf{Inter-stage overlap.}
NPU visual encoding overlaps language execution in both modes. Once a
visual payload is published, the NPU encodes later observations while
the language lanes consume completed inputs. Steady-state throughput
follows the slower of the visual and language paths when they co-run.
Offloading vision frees the GPU for language execution. Successive
single-image requests can also be pipelined across language lanes: one
prefills the next request while another decodes the current one.

\subsection{Versioned Context and Answer Branches}
\label{sec:context-evolution}

A context version identifies the history available to a query.
Publishing $C_t$ marks that history as complete and makes it available
to queries bound to that version. By default, a query binds to the context
version for the most recent observation that arrived before the query.
If that observation, $f_t$, is still being ingested, the query waits
for $C_t$ to be published.

Before query prefill, \name creates a private answer branch rooted at
$C_t$. Query prefill and decoding extend this branch, while new
observations extend the ingestion root to $C_{t+1}$. Publishing a newer
version makes it available
to later queries without changing an active answer's bound prefix or
token positions. \name retains state while an answer or handoff still
uses it.
Versioning lets ingestion proceed without locking the shared history
for the duration of an answer or copying it in full for each response.
When an application feeds an answer back as input, \name adds it to a
later context version before running queries that depend on that
answer. Each answer matches a sequential execution that ingests
its bound history in order and decodes the query on the same engines
with the same model and decoding settings. Section~\ref{sec:eval} uses
this execution as the sequential reference.

\subsection{Cross-Engine State Continuation}
\label{sec:token-state-interfaces}

\noindent\textbf{Continuation boundaries.}
A context handoff transfers the KV state and positions of a
published history, leaving the receiver to prefill and answer the query
on a private branch. \emph{Phase-level state continuation} transfers
that branch after query prefill, so the receiver proceeds directly to
generation.
On M4 Max, Metal~\cite{apple_metal} sends the branch's KV state,
positions, and final prefill logits to the CPU, which samples the first
answer token and continues decoding without repeating prefill.
Metal can then resume ingestion while the CPU finishes the answer.

Keeping query prefill on the faster source lets a receiver with slow
prefill spend its time decoding. A handoff after query prefill moves
less work off the source than a context handoff, but lets the receiver
finish answers sooner and accept more queries. SDEP considers both
effects when choosing a path.

\noindent\textbf{State interface.}
To obtain the history needed for query prefill, a lane uses the state
interface for reuse, import, or reconstruction. Reuse forks state the lane
already holds, sharing computed history while each answer advances its
own suffix. Reuse requires a matching prefix, token positions, model
configuration, and representation. Import transfers compatible KV state
and positional metadata from another backend, as in
Figure~\ref{fig:exec_concurrent}, at a cost that grows with the history.
When compatible state is unavailable, reconstruction
prefills retained visual embeddings and text without rerunning vision,
so every language engine can continue the stream. SDEP compares the preparation
paths available for the same bound history.

For hybrid models such as Qwen3.5~\cite{qwen35}, a
response forks from its bound context version, sharing its attention
KV prefix but keeping its own recurrent state, which cannot be cut back
to a shorter prefix~\cite{marconi}. Each response holds its bound prefix
until completion and owns its generated suffix.

\noindent\textbf{Token interface.}
\label{sec:vision-language-boundary}
Visual tokens cross runtimes at the projector or merger output,
together with model-specific
metadata~\cite{liu2023improvedllava,llava_onevision,Qwen2VL,Qwen2.5-VL}.
\name uses the Token-Aware Buffer Manager (TABM)~\cite{nanomind} to hand
these tokens to multiple language consumers through shared descriptors.
Each consumer binds the payload in place or copies it into storage its
runtime accepts.
The retained visual payload also supplies the inputs for context
reconstruction.
A TABM slot is shared by its consumers and is released after all of
them and any dependent reconstruction finish. Admission bounds requests
in flight, outstanding context updates per stream, and retained state per
lane, and a full token buffer pauses visual production.
Appendix~\ref{app:failures} describes failure handling.

\subsection{State-Driven Execution Planning}
\label{sec:runtime-coordination}
\label{sec:cpu-gpu-parallel}

SDEP compares predicted completion times to choose where each query
phase runs. It also sets how many queries each lane answers.
Where ingestion runs determines where the next context version can be
reused, while the query's path determines the answering work left on
the ingestion lane. SDEP places ingestion on the engine with the lower
projected load from preparing the next version and serving the observed
query workload. SDEP makes these decisions at backend submission and
completion boundaries.

For a query $r$ whose bound context has been published, SDEP estimates
its completion time on a single lane $\ell$ as
\[
T(\ell, r) = W_\ell + P_\ell(r) + S_\ell(r),
\]
where $W_\ell$ is the remaining time of running and queued work,
$P_\ell(r)$ is the time to prepare the bound context on $\ell$ through
the cheapest available path (zero if it is already retained), and
$S_\ell(r)$ is the time to prefill the query and generate the answer.
Completion-time dispatch selects the path predicted to finish first.
A busy lane wins when its reuse or service-time advantage outweighs
the queueing delay. SDEP uses moving
averages of co-running service times to account for interference,
including that from NPU vision. For a split path, query prefill precedes state
transfer, and decoding starts when both the imported branch and the
destination lane are ready. Both queues and the transfer time enter
the estimate through these dependencies, along with each phase's
service cost.

For live streams, SDEP also adjusts how much answering work stays on
the ingestion lane. When its predicted ingestion and answering time
exceeds the budget between query arrivals, SDEP shifts eligible queries
to an idle secondary lane. This \emph{rate-aware capacity} adjustment
counts the time saved on the ingestion lane after subtracting the
export cost. In Figure~\ref{fig:exec_streaming}, query prefill stays on the GPU while
decoding moves to the CPU (Appendix~\ref{app:sdep}).
Section~\ref{ssec:eval:state_dispatch} evaluates these choices, and
Appendix~\ref{app:stateful_routing} describes execution with retained
context replicas.

\subsection{Backend Realization}
\label{sec:backend-realization}

Backend support determines which continuation paths SDEP can use and
how much state preparation they require.
On M4 Max, the ANE or GPU encodes vision. The CPU and Metal backends
of ggml support state continuation in both directions.
CPU-to-GPU handoff uses faster Metal decoding when CPU ingestion keeps
pace with incoming observations. In the other direction, Metal can
hand the CPU a built context or a prefilled answer branch, combining
fast GPU prefill with additional CPU decoding capacity.
The two language backends share a KV representation, so state transfer
requires no layout conversion. Synchronization and compatible buffer
bindings make the transferred state available to the receiver. For
context reuse, the CPU forks a branch at the bound version's exact
prefix length. Metal keeps a separate ingestion root and prefills
queries on private answer branches.

On RK3588 and QCS6490, the NPU encodes vision and the CPU and GPU form the
two language lanes. GPU-to-CPU state continuation lets the GPU extend a
stream's history while the CPU answers from its bound version. Ready
queries can also execute concurrently on both lanes. MNN exports the
GPU state and imports it into the CPU lane. Subsequent GPU updates
leave the receiver's bound prefix unchanged. On
RK3588, SDEP also selects the ingestion engine at runtime, allowing
either lane to build the context for subsequent queries.
Appendix~\ref{app:setup} gives the backend details.

\section{Implementation and Experimental Setup}
\label{sec:impl}
\noindent\textbf{Backend integration.}
We implement \name on Apple M4 Max~\cite{apple_m4max_2024}, Rockchip
RK3588~\cite{rockchip_rk3588_datasheet}, and Qualcomm
QCS6490~\cite{qualcomm_qcs6490_2025}.
NPU inference uses Core~ML or our compiled ANE
backend (Appendix~\ref{app:ane}) on M4 Max, RKNN/RKNPU2~\cite{rknn_toolkit2}
on RK3588, and QNN~\cite{qnn_sdk} on QCS6490. On M4 Max,
ggml~\cite{ggml} runs language inference on the CPU and, through
Metal, on the GPU, and MLX~\cite{mlx2023} encodes
vision on the GPU. MNN~\cite{mnn} runs CPU and OpenCL~\cite{khronos_opencl}
GPU inference on RK3588 and QCS6490.

\noindent\textbf{Multimodal LLMs.}
Our primary model is \emph{Qwen2.5-VL-3B-Instruct}~\cite{Qwen2.5-VL}.
We also evaluate \emph{LLaVA-OneVision} (OneVision)~\cite{llava_onevision},
\emph{FastVLM}~\cite{fastvlm2025}, \emph{Qwen3.5}~\cite{qwen35}, and two 27B
models, \emph{Qwen3.6} and \emph{Qwen3.8}~\cite{qwen36,qwen38}.
Appendix~\ref{app:setup} gives hardware specifications, model sizes,
and the models used on each platform.

\section{Evaluation}
\label{sec:eval}
We evaluate whether \name outperforms the strongest single-engine
baselines, how state sharing contributes to these gains, and why SDEP
must account for differences across architectures and workloads.

\subsection{Methodology}
\label{ssec:metrics}

\noindent\textbf{Baselines and metrics.}
On M4 Max, the external GPU baselines are MLX~\cite{mlx2023} and
llama.cpp~\cite{llama_cpp}. Single-lane runs of \name on the same
backends measure the benefit of a second engine; ablations isolate
context reuse, overlap, state continuation, and SDEP.
Video-MME~\cite{video_mme} provides multi-frame video inputs, and we
replay the timestamped online-video queries of OVO-Bench~\cite{ovo_bench}
as live streams.
\textit{Time to first token} (TTFT) measures the time from request arrival to the
first emitted token, including queueing and all processing.
We measure throughput from the first arrival to the last completion.
End-to-end (E2E) throughput counts generated tokens per second across
all lanes. Request and frame throughput count completed requests and
processed frames per second, respectively.
Measurements start after model loading, compilation, and backend
initialization, and each stage is timed until its output is ready.
Appendix~\ref{app:setup} gives run counts, generation caps, and
statistics.

\subsection{End-to-End Performance}
\label{ssec:eval:e2e}

\noindent\textbf{State-Driven Execution Planning.}
For a 32-frame, 1-fps Qwen2.5-VL-3B stream with 128-token answers on
M4 Max, SDEP cuts p95 TTFT by $76\%$, from $3.56$ to $0.86$\,s,
relative to GPU-only execution with context reuse. Mean TTFT falls
by $67\%$, from $1.61$ to $0.53$\,s, and every query receives its first
token within one second. The p95 full-answer latency falls by $20.0\%$,
and request throughput rises by $9.3\%$. Under backlog, SDEP raises
request throughput by up to $35\%$ over GPU-only
(Figure~\ref{fig:sdep_mechanisms}; Appendix~\ref{app:sdep}).
For the 27B hybrid Qwen3.6 and Qwen3.8 models~\cite{qwen36,qwen38} on an
eight-frame, eight-query backlog, SDEP raises request throughput by up
to $17.5\%$ over GPU-only execution with context reuse. Mean TTFT falls
by up to $22.8\%$.
On two concurrent eight-frame RK3588 streams with NPU visual
encoding, SDEP coordinates both execution modes to outperform the best
single lane with context reuse by $19.6\%$ in request throughput.
Mean and p95 TTFT fall by $34.9\%$ and $36.4\%$, respectively
(Figure~\ref{fig:context_placement}(c,d)).
Against Naive Parallelism, which runs both lanes without shared state,
the throughput gain reaches $27.2\%$
(Appendix~\ref{app:integrated_multistream}).

\begin{figure}[htb]
    \centering
    \includegraphics[width=0.8\columnwidth]{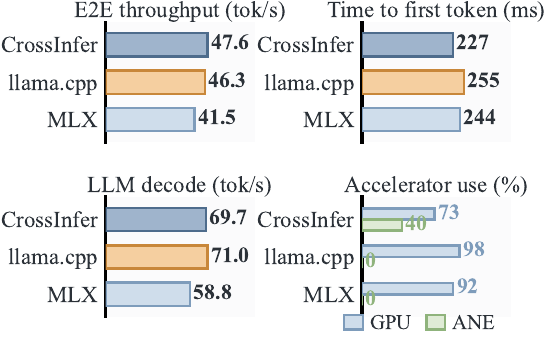}
    \caption{Single-request performance of FP16 Qwen2.5-VL-3B on M4 Max.}
    \label{fig:e2e_teaser}
\end{figure}

\noindent\textbf{Single requests.}
For a single Qwen2.5-VL-3B request on M4 Max, \name uses $23.6\%$ less
CPU, GPU, and ANE energy than llama.cpp (Appendix~\ref{app:setup}). It
reaches the first token in $227$\,ms, versus $244$\,ms for MLX and
$255$\,ms for llama.cpp, and delivers up to $14.9\%$ higher E2E
throughput (Figure~\ref{fig:e2e_teaser}).
Combining text and visual prefill into one pass cuts prefill latency by up to
$28.5\%$ and TTFT by up to $14.2\%$ over three
separate prefills, as in llama.cpp
(Appendix~\ref{app:setup}).

\noindent\textbf{Multi-frame inputs.}
\name pipelines ANE visual encoding with GPU language execution,
reaching $29.6\%$ higher throughput than GPU-only MLX on 64-frame
Video-MME inputs (Appendix~\ref{app:stream_profile},
Table~\ref{tab:video_mme}).
\begin{table}[htb]
    \centering
    \caption{FastVLM-1.5B on Apple M4 Max with context reuse.
    Sequential is \name's path without overlap.
    In (b), all four configurations serve 0.126\,req/s.
    Bold marks better latency and throughput.}
    \label{tab:video_e2e}
    \small
    \setlength{\tabcolsep}{3pt}
    \renewcommand{\arraystretch}{1.1}
    \textbf{(a) FastVLM-1.5B stream}\par\smallskip
    \setlength{\tabcolsep}{4pt}
    \begin{tabular}{@{}lrrr@{}}
        \toprule
        Configuration
        & req/s (SD)$\uparrow$ & TTFT (ms)$\downarrow$ & RSS (GiB) \\
        \midrule
        \multicolumn{4}{@{}l}{\emph{32 tokens per answer}} \\
        GPU-only ggml
        & 3.2780\,{\scriptsize(0.0149)} & 1294 & 2.039 \\
        GPU-only MLX
        & 3.3322\,{\scriptsize(0.0123)} & 1273 & 1.289 \\
        Sequential
        & 3.1044\,{\scriptsize(0.0269)} & 1305 & 1.780 \\
        \name
        & \textbf{4.0976}\,{\scriptsize(0.0365)} & \textbf{994} & 1.826 \\
        \midrule
        \multicolumn{4}{@{}l}{\emph{128 tokens per answer}} \\
        GPU-only ggml
        & 1.3821\,{\scriptsize(0.0069)} & 2757 & 2.052 \\
        GPU-only MLX
        & 1.4347\,{\scriptsize(0.0058)} & 2647 & 1.285 \\
        Sequential
        & 1.3394\,{\scriptsize(0.0057)} & 2787 & 1.842 \\
        \name
        & \textbf{1.5059}\,{\scriptsize(0.0065)} & \textbf{2458} & 1.841 \\
        \bottomrule
    \end{tabular}
    \par\medskip
    \textbf{(b) FastVLM-1.5B, 64 frames at 1\,fps}\par\smallskip
    \setlength{\tabcolsep}{4pt}
    \begin{tabular}{@{}lrrrr@{}}
        \toprule
        & GPU ggml & GPU MLX & Sequential & \name \\
        \midrule
        TTFT (ms)$\downarrow$ & 277 & 302 & 300 & \textbf{245} \\
        \bottomrule
    \end{tabular}
\end{table}

\noindent\textbf{Inter-stage overlap.}
For FastVLM-1.5B, \name
delivers up to $23\%$ higher request throughput and $22\%$ lower mean
TTFT than GPU-only MLX. Overlap alone adds $32.0\%$ throughput over the
same path run sequentially. At both answer lengths, \name also outperforms
GPU-only ggml with $10\%$ less peak resident memory
(Table~\ref{tab:video_e2e}(a)). On a 64-frame stream at 1\,fps, \name
lowers mean TTFT to 245\,ms, $11.7\%$ below GPU-only ggml and $19.1\%$
below GPU-only MLX (Table~\ref{tab:video_e2e}(b)).

For short answers, visual processing takes a larger share of request
time, so overlap yields larger gains. Across four Qwen3.5 models from
0.8B to 9B, throughput gains from overlap reach $72.4\%$ for 32-token
answers and $23.9\%$ for 128-token answers over the same pipeline with
overlap disabled. Session times predicted before the 64-token runs, by
interpolating phase costs and replaying pipeline dependencies, were all
within $1.3\%$ of measurement (Appendix~\ref{app:stream_profile}).
Encoding later requests on the ANE while Metal generates raises
two-stream throughput by $25.7\%$, or $29.1\%$ with more streams, and
lowers median TTFT from $1.75$ to $1.45$\,s
(Figure~\ref{fig:ovo_replay}(a,b)). On a 16-frame Qwen2.5-VL-3B stream,
descriptor handoff with overlap delivers $1.83\times$ the frame
throughput of sequential split execution with byte transfer
(Appendix~\ref{app:stream_profile}).

\subsection{Stateful Streaming Generation}
\label{sec:eval:overlap}

\begin{figure}[tb]
\centering
\includegraphics[width=\columnwidth]{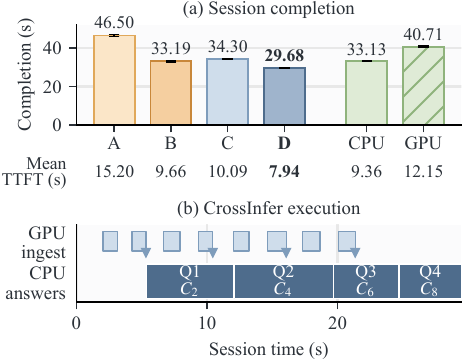}
\caption{Stateful Streaming Generation on RK3588. A--D: neither reuse
nor overlap, overlap only, reuse only, and both (\name).
Error bars show SD; triangles mark context readiness.}
\label{fig:stateful}
\end{figure}

Context reuse avoids rebuilding history, while ingestion overlap lets
the next context update proceed during an answer. We isolate these
effects with fixed roles: the CPU ingests frames and the GPU answers
on M4 Max, with the roles reversed on RK3588
(Figure~\ref{fig:stateful}(b)).
\name delivers $2.48\times$ the request throughput and $70.8\%$ lower
mean TTFT on M4 Max than the same engines run sequentially without context
reuse (Appendix~\ref{app:stateful_details}). Against the same baseline
on an eight-frame, four-query RK3588 stream, throughput rises by
$56.7\%$ and mean TTFT falls by $47.7\%$
(Figure~\ref{fig:stateful}(a)).
Against single-engine execution with context reuse, \name also raises
request throughput by $7.9\%$ over GPU-only on M4 Max, and by $11.6\%$
and $37.2\%$ over CPU-only and GPU-only on RK3588.

With context reuse in both runs, ingestion overlap adds $12.2\%$
throughput on M4 Max and $15.6\%$ on RK3588.
Reusing the visual embeddings and shared prefix across four queries delivers
$2.34\times$ the throughput of recomputing the prefix on M4 Max
and $2.79\times$ on RK3588 (Appendix~\ref{app:context_results}).
For larger MLLMs on the RK3588 CPU, the gains reach $2.89\times$ for
Qwen2.5-VL-3B and $1.90\times$ for OneVision-7B, whose later queries
start answering in about 1\,s.

With GPU ingestion and CPU answering, this mode
outperforms the best single lane for every model on RK3588 and
QCS6490, by up to $22.4\%$ (Figure~\ref{fig:multimodel_strategies}).
The gain over GPU-only execution with context reuse reaches $59.4\%$
on QCS6490.
For OneVision-7B, ingestion overlap
cuts the second query's TTFT by $11.0$\,s versus CPU-only execution.
On QCS6490, ingestion overlap adds
$20.8$--$24.3\%$ throughput over the same engines run sequentially with
context reuse (Appendices~\ref{app:multimodel_strategies} and~\ref{app:qcs6490}).

\subsection{Concurrent Sequence Generation}
\label{ssec:eval:concurrent}

\begin{figure}[htb]
\centering
\subfigure[M4 Max: throughput]{
\includegraphics[width=0.455\columnwidth]{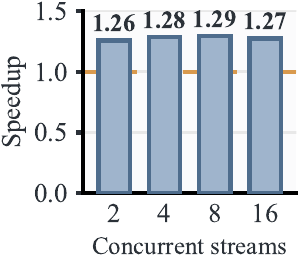}
\label{fig:ovo_throughput}
}%
\subfigure[M4 Max: median TTFT]{
\includegraphics[width=0.455\columnwidth]{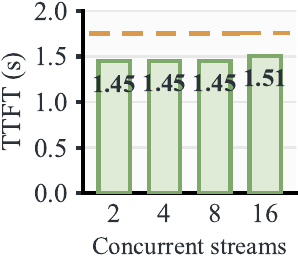}
\label{fig:ovo_ttft}
}\\
\subfigure[RK3588: throughput]{
\includegraphics[width=0.455\columnwidth]{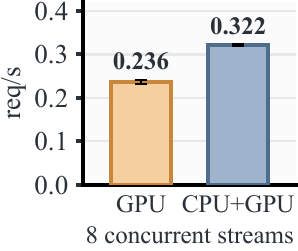}
\label{fig:ovo_throughput_rk}
}%
\subfigure[RK3588: p95 latency]{
\includegraphics[width=0.455\columnwidth]{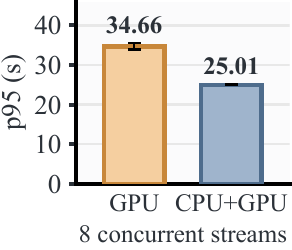}
\label{fig:ovo_ttft_rk}
}
\caption{(a,b) M4 Max overlaps ANE vision with one Metal language lane;
dashed lines mark the same path serving one request at a time.
(c,d) Concurrent Sequence Generation on RK3588 versus GPU-only at
eight streams; error bars show SD.
}
\label{fig:ovo_replay}
\end{figure}

\begin{figure}[!t]
\centering
\subcapraggedrighttrue
\subfigure[M4 Max: CPU + Metal GPU]{%
\includegraphics{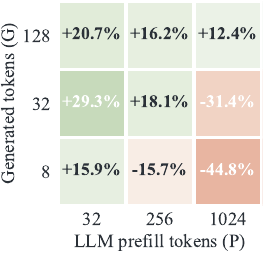}%
\label{fig:lane_gain_mac}%
}\hfill
\subfigure[RK3588: \mbox{CPU + Mali GPU}]{%
\includegraphics{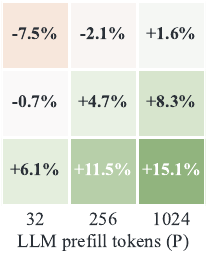}%
\label{fig:lane_gain_map}%
}\\
\subfigure[Co-running decode]{%
\includegraphics{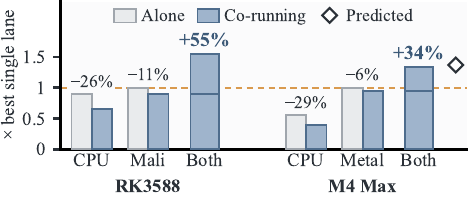}%
\label{fig:lane_interference}%
}\\
\subfigure[Two modes across models]{%
\includegraphics{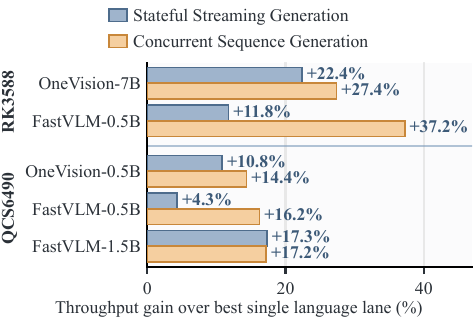}%
\label{fig:multimodel_strategies}%
}
\caption{(a,b,d) Gains over the best single lane. (c) Slowdowns and the
M4 Max prediction.
(d) RK3588 uses precomputed visual embeddings; QCS6490 includes
on-device vision.}
\label{fig:lane_boundary}
\end{figure}

On RK3588, the CPU and GPU lanes deliver $36.3\%$ higher request
throughput than GPU-only at eight concurrent streams and cut p95
latency by $27.8\%$
(Figure~\ref{fig:ovo_replay}(c,d);
Appendix~\ref{app:concurrent_sweep}). In a concurrency sweep from two to
sixteen streams, the throughput gain reaches $52.0\%$. With four simultaneous requests,
throughput exceeds the best single lane by $27.4\%$ for
OneVision-7B and $37.2\%$ for FastVLM-0.5B on RK3588, and by
$14.4$--$17.2\%$ for all three models on QCS6490 with on-device vision
(Figure~\ref{fig:multimodel_strategies}).

\begin{figure}[t]
\centering
\subfigure[Live-stream latency]{\includegraphics[width=0.48\columnwidth]{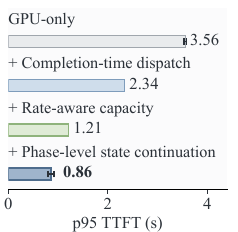}}\hfill
\subfigure[Throughput under backlog]{\includegraphics[width=0.48\columnwidth]{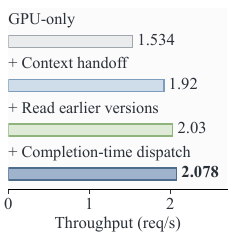}}
\caption{SDEP mechanisms added cumulatively on M4 Max, with context
reuse. Whiskers show SD.}
\label{fig:sdep_mechanisms}
\end{figure}

\subsection{When Two Language Lanes Improve Throughput}
\label{ssec:eval:lane_boundary}

\begin{figure}[!t]
\centering
\subfigure[Decoding]{\includegraphics[width=0.32\columnwidth]{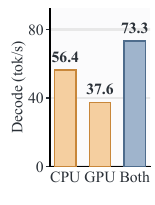}\label{fig:rk_ablation_decoding}}\hfill
\subfigure[Buffer handoff]{\includegraphics[width=0.32\columnwidth]{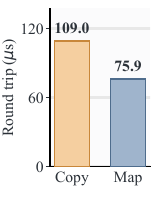}\label{fig:rk_ablation_ttft}}\hfill
\subfigure[KV import]{\includegraphics[width=0.32\columnwidth]{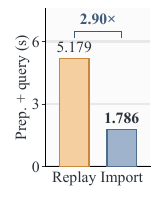}\label{fig:kv_import}}
\caption{RK3588 mechanisms. (a) CPU/GPU decoding of
Qwen2.5-0.5B~\cite{qwen25}. (b) Buffer mapping versus copying.
(c) OneVision-0.5B preparation and query time with GPU KV import
versus CPU replay.}
\label{fig:rk_ablation}
\end{figure}

Whether a second language lane improves throughput depends on the
relative cost of prefill and decoding. M4 Max and RK3588 show opposite
trends.
On M4 Max, adding the CPU lane improves throughput in six of nine
configurations, by up to $29.3\%$ with short prefill and moderate
generation, but reduces it by $44.8\%$ at $1024$ prefill tokens and
eight output tokens (Figure~\ref{fig:lane_boundary}). For Qwen2.5-VL-3B,
Metal prefills $8.0\times$ faster than the CPU but decodes only
$1.7\times$ faster. With the 1024-token history already retained,
the CPU's decoding contribution offsets Metal's co-running slowdown,
raising aggregate throughput by $33.9\%$ over Metal alone. This rate is
within $2.3\%$ of the pre-run prediction
(Figure~\ref{fig:lane_interference}; Appendix~\ref{app:mac_lane}).
On RK3588, where the lanes run at closer rates, gains grow with prefill
length and shrink as longer generation increases shared-memory
contention through repeated weight reads. With 256 generated tokens,
adding the GPU lane raises decode throughput to $73.3$\,tok/s,
$30\%$ above the faster CPU lane's $56.4$\,tok/s
(Figure~\ref{fig:rk_ablation_decoding}).
Which lane is faster also depends on the model
(Appendix~\ref{app:multimodel_strategies}).

\subsection{State-Driven Execution Planning}
\label{ssec:eval:state_dispatch}

On the live M4 Max stream, dispatch by predicted completion time sends
queries to the CPU using Metal-built context, lowering p95 TTFT from
$3.56$ to $2.34$\,s. Rate-aware capacity then adjusts the CPU's share
of queries, reducing p95 TTFT to $1.21$\,s. Phase-level state continuation lets
Metal prefill queries assigned to the CPU, freeing the CPU to spend
its answering time decoding. This reduces p95 TTFT by a further
$28.9\%$, to $0.86$\,s, and mean TTFT by $10.6\%$
(Figure~\ref{fig:sdep_mechanisms}(a)). Under backlog, SDEP uses context
handoff to let the CPU answer from its bound version while ingestion
advances, reaching $2.078$\,req/s versus $1.534$\,req/s with GPU-only execution
(Figure~\ref{fig:sdep_mechanisms}(b)).

\begin{figure}[!t]
\centering
\subcapraggedrighttrue
\includegraphics{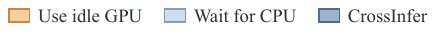}\\
\subfigure[Reuse vs.\ idle capacity]{%
\includegraphics[height=121.4bp]{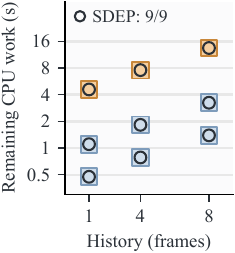}%
\label{fig:context_placement_choice}%
}\hfill
\subfigure[Query completion]{%
\includegraphics[height=121.4bp]{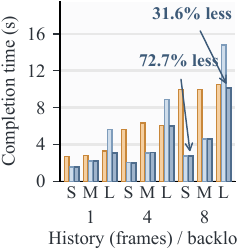}%
\label{fig:context_placement_completion}%
}\\
\subfigure[Throughput]{%
\includegraphics[height=91bp]{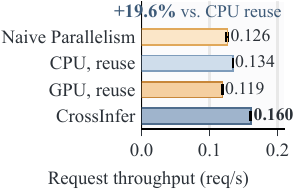}%
\label{fig:context_placement_throughput}%
}\hfill
\subfigure[Mean TTFT]{%
\includegraphics[height=91bp]{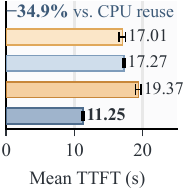}%
\label{fig:context_placement_ttft}%
}
\caption{SDEP on RK3588. (a) Colors mark the faster fixed
policy; circles mark SDEP choices. (b) S/M/L denote short/medium/long
backlogs. (c,d) Two concurrent streams; error bars show SD.}
\label{fig:context_placement}
\end{figure}

On RK3588, SDEP selects the faster lane in all nine history/backlog
settings (Figure~\ref{fig:context_placement}(a,b)). Longer histories
increase the value of reuse, but a longer queue can make the idle GPU
faster. At eight frames, waiting for CPU reuse cuts query completion
time by $72.7\%$ over using the idle GPU under a short backlog.
Under a long backlog, the choice reverses: using the GPU cuts completion
time by $31.6\%$ over waiting for the CPU. On two QCS6490 streams,
SDEP raises request throughput by $43.7\%$ over prioritizing context
reuse and $7.3\%$ over prioritizing idle lanes
(Appendix~\ref{app:context_placement}). It reduces mean TTFT by
$9.0\%$ relative to CPU-only with context reuse, or by
$17.9\%$ under sparse arrivals.

\subsection{Mechanism Analysis}
\label{ssec:eval:ablation}

On RK3588, importing GPU context instead of recomputing the prefix
cuts first-token latency for CPU answers by a factor of $4.6$,
from $1482$ to $319$\,ms, and serves three queries $2.90\times$ faster
(Figure~\ref{fig:kv_import}; Appendix~\ref{app:state_handoff}).
Reconstructing upcoming contexts on the GPU while the CPU answers
cuts a later query's time to first token by $41.0\%$
relative to CPU-only execution (Appendix~\ref{app:stateful_routing}).
On M4 Max, descriptor handoff of visual tokens avoids serialization
and copying, cutting the transport round trip by a factor of $12.1$ and delivering
$3.3\times$ the frame throughput under backlog (Appendix~\ref{app:lane_price}),
while on RK3588, mapping buffers instead of copying them cuts the
round trip by $30\%$ (Figure~\ref{fig:rk_ablation_ttft}).

\subsection{Output Equivalence and Accuracy}
\label{sec:eval:correctness}

Across all three platforms, concurrent and overlapped runs match
their sequential references in output tokens and first-step logits for
every answer, including all 96 M4 Max answers using phase-level state
continuation (Appendix~\ref{app:sdep}).
Sequential and overlapped runs of OneVision-7B and FastVLM-0.5B also
produce bit-identical per-layer K/V tensors
(Appendix~\ref{app:context_results}). CPU-to-Metal handoff matches
CPU-local accuracy at $75.0\%$ on M4 Max, while GPU-to-CPU handoff
achieves comparable accuracy for all three models on QCS6490
(Table~\ref{tab:correctness}).

\begin{table}[!t]
\caption{Task accuracy (\%) after state handoff on 200 single-frame
OVO queries per model, with 4-bit language weights.
(a) Qwen2.5-VL-3B on M4 Max. (b) QCS6490.}
\label{tab:correctness}
\centering
\small
\setlength{\tabcolsep}{4pt}
\renewcommand{\arraystretch}{1.0}
\textbf{(a) M4 Max: CPU-to-Metal handoff}\par\smallskip
\begin{tabular}{lccc}
\toprule
& CPU local & Metal local & CPU $\rightarrow$ Metal \\
\midrule
Accuracy (\%) & 75.0 & 74.5 & \textbf{75.0} \\
\bottomrule
\end{tabular}
\par\medskip
\textbf{(b) QCS6490: GPU-to-CPU handoff}\par\smallskip
\begin{tabular}{lcc}
\toprule
Model & CPU local & GPU $\rightarrow$ CPU \\
\midrule
OneVision-0.5B & 47.0 & 50.5 \\
FastVLM-0.5B   & 69.5 & 70.5 \\
FastVLM-1.5B   & 74.0 & 72.5 \\
\bottomrule
\end{tabular}
\end{table}

\section{Related Work}
\label{sec:related}
\subsection{Heterogeneous On-Device Inference}
NanoMind~\cite{nanomind} decomposes multimodal inference into modules
placed across processors and connects them through TABM, and Phase
Matters~\cite{phasematters} characterizes heterogeneous MLLM pipeline
configurations on a mobile SoC. llm.npu~\cite{llmnpu},
PowerInfer-2~\cite{powerinfer2}, HeteroInfer~\cite{heteroinfer},
HeteroMosaic~\cite{heteromosaic}, and pipelined
sharding~\cite{xlmclients} split LLM and MLLM inference across CPUs, GPUs,
and NPUs, while Agent.xpu~\cite{agentxpu} schedules concurrent agentic
workloads and CORE~\cite{coredvfs} coordinates CPU, GPU, and memory
frequencies. Core
ML~\cite{apple_coreml} and ExecuTorch~\cite{executorch} map one model's
operations onto several backends within one framework. \name coordinates
context updates and query branches throughout a serving session,
carrying visual tokens and language state between engines and continuing
queries across prefill and decoding.

\subsection{Prefill--Decode Scheduling and Disaggregation}
Orca~\cite{orca} introduces iteration-level request scheduling,
while vLLM~\cite{vllm} uses PagedAttention for memory-efficient
request batching.
DistServe~\cite{distserve} and Splitwise~\cite{splitwise} separate
prefill and decoding across GPUs or machines in serving clusters;
Encode--Prefill--Decode (EPD)~\cite{epd} and TriInfer~\cite{triinfer}
also disaggregate multimodal encoding.
Sarathi-Serve~\cite{sarathi} and NanoFlow~\cite{nanoflow} chunk or
overlap work within and across stages, and Stream2LLM~\cite{stream2llm}
overlaps prefill with streaming context retrieval. Preble~\cite{preble} balances KV reuse and load in distributed
serving, and performance models choose CPU--GPU policies per
phase~\cite{flexinfer} or predict co-running slowdowns on edge
devices~\cite{pitot}. Within one SoC, SDEP decides per query whether to
split prefill and decoding, transferring the answer branch through
unified memory while preserving its bound context version.

\subsection{Context Reuse and State Transfer}
Prompt Cache~\cite{promptcache} and SGLang~\cite{sglang} reuse attention
states and KV prefixes across requests, while LMCache~\cite{lmcache} and
SuperInfer~\cite{superinfer} move KV between engines or between GPU and
CPU memory. On devices,
SparKV~\cite{sparkv} overlaps cloud-to-device KV loading
with local recomputation. Dynamic Flow, Static Graph~\cite{dynamicflow}
maps prefix and non-prefix reuse onto statically compiled mobile NPU
graphs. mzCache~\cite{mzcache} restores evicted weights and KV through
shared CPU--GPU buffers while GPU inference proceeds. \name reuses
context across backends with different KV representations, importing
compatible state or reconstructing the context from retained visual
embeddings and text. Each answer branch retains its bound history
while ingestion advances the stream.

\subsection{Streaming Video Understanding}
VLMCache~\cite{vlmcache} reuses visual computation across frames
to reduce on-device prefill cost, and ReKV~\cite{rekv} decouples video
processing from question answering through video KV caches.
InfiniPot-V~\cite{infinipotv}, StreamingVLM~\cite{streamingvlm}, and
V-Rex~\cite{vrex} keep the growing video KV cache within budget by
compression, by retaining attention sinks and recent tokens, or by
dynamic retrieval on a dedicated accelerator.
Speak While Watching~\cite{speakwhilewatching}, Think While
Watching~\cite{thinkwhilewatching}, and Think-as-You-See~\cite{tays}
support concurrent visual ingestion
and generation through changes to positional encoding, attention,
and state management. \name runs ingestion and generation on separate
engines with the model unchanged, binding each response to a published
version.

\section{Conclusion}
\label{sec:con}
\name makes engine roles a runtime choice for stateful MLLM inference.
Versioned state continuation lets engines take over after context
ingestion or query prefill. For the same workload, a second language
engine can raise throughput on one SoC and lower it on another. Its
benefit depends on the work it takes over, the state it needs, and the
interference it introduces. SDEP uses these costs and arrival rates to
decide where each query phase runs and how many queries each engine
answers, coordinating ongoing ingestion and answering.

\bibliographystyle{mlsys2025}
\bibliography{reference}

\section*{AI Usage Disclosure}
OpenAI Codex and Anthropic Claude Code were used iteratively to generate and
revise runtime code, experimental scripts.
They also supported testing, evaluation, and result analysis. The authors directed these
iterations.

\clearpage
\appendix
\section{Additional ANE Backend and CRANE Compilation}
\label{app:ane}

Alongside the Core~ML~\cite{apple_coreml} visual path, our custom backend
invokes Apple's private ANE~\cite{apple_ane_report} interfaces. Our CRANE
pipeline specializes visual graphs for static compilation.

\subsection{Runtime and Static Graph Compilation}
\label{app:ane:runtime}

Open-source ANE work~\cite{ane_maderix} exposes private compilation
and execution interfaces, including fused computation and shared
IOSurface buffers~\cite{apple_iosurface}. Orion~\cite{orion} adds program caching, graph
compilation, and inference support. Our custom visual backend uses
these private interfaces to submit Model Intermediate
Language~(MIL) programs~\cite{apple_mil} to Apple's compiler and execute the
resulting programs on the ANE.

Compiling a visual encoder for an NPU requires supported operators
and tensor shapes, tiling, and positional schedules compatible with
the backend's compiler.
CRANE separates host-side schedule construction from static
device computation. For a fixed image geometry, the host determines
the patch grid, window indices, and positional metadata before
compilation. Specializing these values produces a static compute
graph while retaining the model's attention and positional
semantics. Different image geometries require corresponding
specializations. The same separation also makes fixed-schedule
visual computation suitable for Core~ML export.

\subsection{Fused Execution and Buffer Chaining}
\label{app:ane:fusion}

The fixed-resolution ANE backend compiles each vision transformer
block into one MIL program. Each program fuses normalization,
projections, attention, the feed-forward network, and residual
updates. This reduces host work between operators. Model weights
and fixed rotary-position tensors are embedded as compile-time
constants, and compiled programs are reused across requests with
the same input geometry.

Blocks pass intermediate activations through IOSurface-backed
buffers. A cached execution plan reuses these buffers and avoids
repeated host packing between blocks. Chaining keeps intermediate
activations within the visual encoder. The evaluated
\texttt{ane\_fused} path returns the final output to the host;
the language backend then copies it into its input tensor.
Descriptor handoff and payload copying remain separate operations.

\subsection{Compilation Granularity and Numerical Checks}
\label{app:ane:checks}

On the fixed $384{\times}384$ Qwen2.5-VL-3B~\cite{Qwen2.5-VL} configuration,
the fused and chained visual path of the compiled backend reaches a
warm visual-encoding latency of $245$\,ms across $32$ transformer
blocks. A single block with
approximately $39$\,MB of embedded weights and constants compiles
into one MIL program. RKNN's whole-encoder submission~\cite{rknn_toolkit2} and this backend's
block-level submission are different ways of realizing the
visual stage.

For the evaluated pooler outputs, fused versus unfused ANE
execution has a maximum absolute difference of $0.079$ and a mean
of $0.003$. Comparing fused ANE with the bfloat16 MLX~\cite{mlx2023} path gives
a maximum difference of $0.066$ and a mean of $0.003$.
Results labeled \texttt{ane\_fused} use this custom backend.

\section{Experimental Setup Details}
\label{app:setup}

\noindent\textbf{Hardware platforms.}
The Mac Studio~\cite{apple_mac_studio_2025} uses an Apple M4 Max
SoC~\cite{apple_m4max_2024} with a 16-core Apple Neural
Engine~\cite{apple_ane_report}, a 40-core GPU, and 64\,GB of unified
memory. The Orange Pi 5 Ultra~\cite{orangepi5ultra} uses a Rockchip
RK3588 SoC~\cite{rockchip_rk3588_datasheet} with four
Cortex-A76~\cite{arm_cortex_a76} and four Cortex-A55~\cite{arm_cortex_a55}
CPU cores, a Mali-G610 GPU~\cite{arm_mali_g610}, and a 6\,TOPS NPU.
The Rubik Pi 3~\cite{rubikpi3,qualcomm_rubikpi3} uses a Qualcomm
QCS6490 SoC~\cite{qualcomm_qcs6490_2025} with four
Cortex-A78~\cite{arm_cortex_a78} and four Cortex-A55 CPU cores, an
Adreno 643 GPU~\cite{qualcomm_adreno}, and a 12\,TOPS Hexagon
NPU~\cite{qualcomm_hexagon}.

\noindent\textbf{Models and platform coverage.}
Qwen2.5-VL-3B-Instruct~\cite{Qwen2.5-VL} is the primary model.
The additional models are LLaVA-OneVision (OneVision)~\cite{llava_onevision},
built on Qwen2~\cite{qwen2}, at 0.5B and 7B;
FastVLM~\cite{fastvlm2025} at 0.5B and 1.5B; and
Qwen3.5~\cite{qwen35} at 0.8B, 2B, 4B, and 9B.
OneVision-0.5B is used for the main RK3588 streaming, context-reuse,
and State-Driven Execution Planning experiments.
OneVision-7B and FastVLM-0.5B run both execution modes on RK3588;
OneVision-0.5B and both FastVLM models run them on QCS6490.
We use Qwen3.5 to evaluate inter-stage overlap across model sizes on M4 Max.

\noindent\textbf{Engine backends.}
On RK3588, RKNN-Toolkit2~\cite{rknn_toolkit2} compiles the visual encoder
and projector into a fixed-shape \texttt{.rknn} graph, which RKNPU2
submits as one execution unit. In Concurrent Sequence Generation, the
OpenCL~\cite{khronos_opencl} worker prepares context for each request it
serves. On Apple M4 Max, when ingestion and generation share a ggml~\cite{ggml} KV representation,
continuation needs synchronization and compatible buffer bindings rather
than layout conversion. On every platform, image preprocessing and
model-specific input preparation stay on the host, compilation and
reusable-buffer setup happen outside the per-request path, and each
backend reports completion before TABM~\cite{nanomind} publishes its output.

\begin{figure}[htb]
    \centering
    \includegraphics[width=0.8\columnwidth]{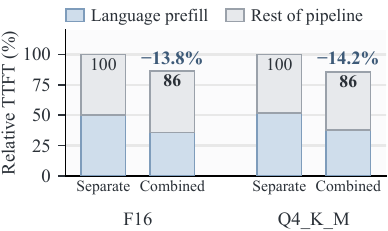}
    \caption{Single-request TTFT for Qwen2.5-VL-3B on M4 Max with separate and combined prefill (\name). Vision runs through Core~ML on the ANE, language through ggml on Metal.}
    \label{fig:single_request_prefill}
\end{figure}

On M4 Max, the streaming comparison reports the mean of five runs per configuration, with up to 32 generated tokens per request and 16 frames per run.
On RK3588, the concurrency sweep runs
a 16-request stream per configuration on the CPU and
OpenCL lanes with an eight-token generation cap
and reports aggregate request throughput and p50/p95 end-to-end latency.
Figure~\ref{fig:e2e_teaser} uses Qwen2.5-VL-3B in FP16, with
Core~ML visual encoding on the ANE and ggml~\cite{ggml} language
inference on Metal~\cite{apple_metal} with combined prefill. All three
systems answer the same image and prompt. We report the mean of five
measured runs.
TTFT runs from the image request to the first token; for \name it
comprises $4.5$\,ms of preprocessing, $121.3$\,ms of visual encoding,
$100.5$\,ms of language prefill, and $1.0$\,ms of other first-token work.
GPU use is the mean device-level sample over the request. For 32 output
tokens, E2E throughput is
$32/(\mathrm{TTFT}+31/r)$, where TTFT is in seconds and $r$ is the
decode rate in tok/s. Each request starts from fresh KV state with
prefix reuse disabled. Table~\ref{tab:mac_energy} gives per-request
energy for \name, llama.cpp~\cite{llama_cpp}, and MLX over five 32-token
requests each, with CPU, GPU, and ANE power from
powermetrics~\cite{apple_powermetrics}.
Figure~\ref{fig:single_request_prefill} compares separate and combined
prefill on the same path in F16 and Q4\_K\_M. Each configuration runs
five warm requests that generate one token, and TTFT runs from the image
request to the first token. Separate prefill processes the prompt prefix,
the visual rows, and the prompt suffix in three passes; combined prefill
processes them in one pass over the same handoff buffer.
Over two questions with up to 32 generated tokens, combined prefill matches
separate prefill token for token in Q4\_K\_M; in F16, one answer diverges
at token 12, where the two leading candidates differ by $0.0003$ in logit.

\begin{table}[tb]
\centering
\caption{Per-request energy and mean power on M4 Max for Qwen2.5-VL-3B
in FP16 with 32 output tokens.}
\label{tab:mac_energy}
\small
\begin{tabular}{lrrr}
\toprule
System & GPU (J) & CPU+GPU+ANE (J) & Power (W) \\
\midrule
\name     & 18.57 & \textbf{20.74} & \textbf{29.49} \\
llama.cpp & 25.80 & 27.13 & 38.94 \\
MLX       & 18.50 & 21.52 & 30.34 \\
\bottomrule
\end{tabular}
\end{table}

In the five-run matched eight-stream OVO~\cite{ovo_bench} comparison
in Section~\ref{ssec:eval:concurrent}, each request carries roughly 730
visual tokens and generates at most eight tokens. Requests are dispatched to idle language lanes using observed service times.
Shared-prefix reuse and eight-frame stateful execution also use five
complete runs per configuration. Error bars and $\pm$ values denote
sample standard deviations across runs; percentage-gain deviations
are in percentage points. Gains are calculated within each paired run
and then averaged. At eight concurrent streams, p95 is measured from
batch release to request completion.

\noindent\textbf{Visual inputs.}
The RK3588 streaming and placement experiments with
OneVision~\cite{llava_onevision} pool the projected spatial grid from $27\times27$ to $14\times14$, retaining
one newline row for 197 visual rows per frame. The eight-stream OVO
replay uses the original roughly 730-row image representation.
For OneVision-0.5B, reducing the grid cuts median OpenCL prefill time
by $54.0\%$ and raises request throughput by $56.1\%$. This comparison
varies the visual input; execution-mode comparisons keep it fixed.
FastVLM-0.5B~\cite{fastvlm2025} uses its own 256-row representation
(Appendix~\ref{app:multimodel_strategies}).

Visual features enter the language model according to its architecture:
cross-attention models~\cite{flamingo} connect them to attention layers,
and Qwen3-VL's DeepStack~\cite{Qwen3-VL} additionally injects them into
intermediate language layers.

\noindent\textbf{Language precision.}
INT8 gives the highest accuracy in the 200-query OneVision-0.5B
OpenCL evaluation (Table~\ref{tab:language_precision}). Each query
uses one OVO frame and one generated choice token, with the same
visual inputs across weight precisions. In a separate CPU context-reuse
comparison, INT4 gives the highest request throughput at
$1.044$\,req/s. That comparison uses four distinct questions,
32-token caps, and a shared 197-row visual prefix.

\begin{table}[tb]
\centering
\caption{OneVision-0.5B language-weight precision on RK3588 OpenCL,
using the same 200 single-frame OVO queries.}
\label{tab:language_precision}
\small
\begin{tabular}{lrr}
\toprule
Weights & Correct / 200 & Accuracy (\%) \\
\midrule
FP16 & 117 & 58.5 \\
INT8 & \textbf{123} & \textbf{61.5} \\
INT4 & 106 & 53.0 \\
\bottomrule
\end{tabular}
\end{table}

\section{Fixed-Resolution Streaming Configuration}
\label{app:stream_profile}

\noindent\textbf{Video-MME inputs.}
Table~\ref{tab:video_mme} compares six Video-MME~\cite{video_mme} input
sizes on M4 Max. \name completes inputs faster than MLX from 16 frames
onward, reaching $29.6\%$ higher throughput at 64 frames
($16.60$ versus $21.51$\,s). Across input sizes, ANE utilization rises
from $27\%$ to $91\%$ and GPU utilization falls from $80\%$ to $13\%$,
while MLX keeps the GPU saturated.

\begin{table}[tb]
\centering
\caption{Video-MME on M4 Max with 150 generated tokens per run.
Input is frames @ fps; bold marks better time and throughput.}
\label{tab:video_mme}
\small
\setlength{\tabcolsep}{3pt}
\renewcommand{\arraystretch}{1.1}
\begin{tabular}{@{}lrrrrrrr@{}}
    \toprule
    & \multicolumn{3}{c}{MLX}
    & \multicolumn{4}{c}{\name} \\
    \cmidrule(lr){2-4}
    \cmidrule(lr){5-8}
    Input
    & Time$\downarrow$ & E2E$\uparrow$ & GPU
    & Time$\downarrow$ & E2E$\uparrow$ & GPU & ANE \\
    & (s) & (tok/s) & (\%)
    & (s) & (tok/s) & (\%) & (\%) \\
    \midrule
    2 @ 0.5
    & \textbf{2.21} & \textbf{67.87} & 85
    & 2.34 & 64.10 & 80 & 27 \\

    4 @ 1
    & \textbf{2.85} & \textbf{52.63} & 89
    & 3.23 & 46.44 & 70 & 43 \\

    8 @ 2
    & \textbf{3.86} & \textbf{38.86} & 93
    & 4.19 & 35.80 & 55 & 56 \\

    16 @ 4
    & 6.33 & 23.70 & 93
    & \textbf{6.10} & \textbf{24.59} & 37 & 76 \\

    32 @ 4
    & 10.96 & 13.69 & 98
    & \textbf{9.94} & \textbf{15.09} & 23 & 85 \\

    64 @ 8
    & 21.51 & 6.97 & 99
    & \textbf{16.60} & \textbf{9.04} & 13 & 91 \\
    \bottomrule
\end{tabular}
\end{table}

The 16-frame comparison in Section~\ref{ssec:eval:e2e} uses the
compiled ANE visual backend (Appendix~\ref{app:ane}) and an FP16
Qwen2.5-VL-3B language model through the native ggml bridge on Metal.
Both modes use these backends. Sequential split execution transfers
visual tensors as bytes and finishes language generation before
encoding the next frame. Overlapped execution uses TABM descriptors
and encodes subsequent frames while the language worker generates
answers. Frame throughput includes token handoff and completed
generation, with up to 32 output tokens per request. Overlap raises
frame throughput from $1.57$ to $2.89$\,fps, $1.83\times$ sequential split
execution, at an unchanged decode rate.

The FastVLM-1.5B comparison in Table~\ref{tab:video_e2e}(a) streams
eight OVO~\cite{ovo_bench} frames with eight questions and generates 32
or 128 tokens per answer. The GPU-only configurations run MLX vision,
encoding the next frame ahead, with 4-bit ggml or MLX language models;
\name and Sequential run Core~ML vision whose output the ggml language model
reads in place, and only \name overlaps it with generation. Every
configuration retains the video KV state and runs five times.
In Table~\ref{tab:video_e2e}(b), each query generates 32 tokens, and each
configuration runs three times.

The Qwen3.5~\cite{qwen35} scale comparison in Section~\ref{ssec:eval:e2e}
runs the same pipeline at 0.8B, 2B, 4B, and 9B: Core~ML vision, Metal
language execution with Q4\_K\_M weights and retained context, and 32-,
64-, or 128-token
answers, sequentially or with inter-stage overlap, three runs per
configuration. Table~\ref{tab:qwen35_scale} gives the request throughput
of sequential and overlapped execution and the gain from overlap at 32 and
128 tokens. At 64 tokens, overlap raises request throughput by
$36.9\%$, $41.7\%$, $22.1\%$, and $17.5\%$ for the 0.8B, 2B, 4B, and 9B
models. Calibrated phase-cost interpolation with dependency-aware pipeline
replay, frozen before these runs, predicts all eight session times, four
models in both modes, within $1.3\%$. It calibrates each stage's cost from
the 32- and 128-token runs, interpolates decoding time linearly in answer
length, and replays the visual--language dependencies of sequential and
overlapped execution. Gains are largest for short answers,
where visual processing takes the largest share of each request. All
288 pairs of sequential and overlapped
answers are identical across the 576 requests. Peak process RSS is
sampled during overlapped execution in the 32- and 128-token runs.

\begin{table*}[t]
\caption{Inter-stage overlap across Qwen3.5 model sizes on Apple M4 Max:
request throughput (req/s, three runs per configuration) of the same
retained-context pipeline run sequentially and with overlap, the gain from
overlap, and peak process RSS during overlapped execution.}
\label{tab:qwen35_scale}
\centering
\small
\setlength{\tabcolsep}{5pt}
\begin{tabular}{lrrrrrrr}
\toprule
& \multicolumn{3}{c}{32-token answers} & \multicolumn{3}{c}{128-token answers} & Peak RSS \\
\cmidrule(lr){2-4} \cmidrule(lr){5-7}
Model & Sequential & Overlap & Gain & Sequential & Overlap & Gain & (GiB) \\
\midrule
Qwen3.5-0.8B & 3.618 & 6.009 & 66.1\% & 1.614 & 1.906 & 18.1\% & 1.261 \\
Qwen3.5-2B   & 2.278 & 3.927 & 72.4\% & 1.069 & 1.324 & 23.9\% & 1.917 \\
Qwen3.5-4B   & 1.434 & 1.963 & 36.9\% & 0.603 & 0.682 & 13.1\% & 3.394 \\
Qwen3.5-9B   & 1.010 & 1.304 & 29.1\% & 0.417 & 0.486 & 16.6\% & 6.094 \\
\bottomrule
\end{tabular}
\end{table*}

\section{Second-Lane Throughput and Buffer Handoff}
\label{app:lane_price}

We run 16 ordered TextVQA~\cite{textvqa} requests using
OneVision-0.5B~\cite{llava_onevision} with a Qwen2~\cite{qwen2} language
backbone, NPU vision, and at most eight generated tokens per request.
Running both language lanes alongside the NPU raises request throughput
by $14.3\%$ over the best single lane, from $0.2774$ to
$0.3172$\,req/s.

\begin{figure}[htb]
\centering
\subfigure[Makespan and response latency]{
\includegraphics[width=0.95\columnwidth]{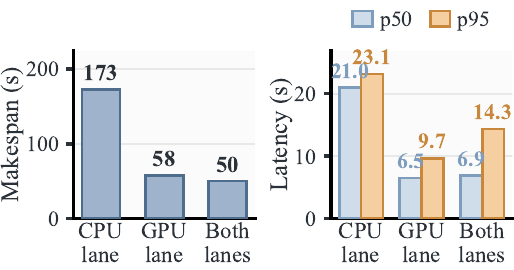}
\label{fig:stream_tradeoff}
}\\
\subfigure[Buffer handoff latency]{
\includegraphics[width=0.9\columnwidth]{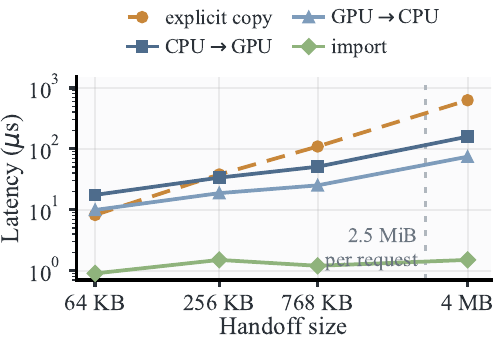}
\label{fig:handoff_cost}
}
\caption{Second-lane throughput and buffer handoff on RK3588. (a) Makespan and p50/p95 response latency over a 16-request stream. (b) Buffer handoff latency versus transfer size, medians of 50 iterations. The vertical line marks the $2.5$\,MiB that \name moves per request.}
\label{fig:lane_cost}
\end{figure}

Adding the CPU lane reduces the 16-request stream's makespan from
$57.7$ to $50.4$\,s, a $12.5\%$ reduction
(Figure~\ref{fig:lane_cost}).

The RK3588 co-running decode measurement in Figure~\ref{fig:lane_interference}
uses 32 prefill tokens and 128 generated tokens. Co-running slows CPU
decoding by $26\%$ and Mali GPU~\cite{arm_mali_g610} decoding by $11\%$,
while aggregate throughput exceeds the Mali GPU single-lane baseline
by $55\%$.
Figure~\ref{fig:rk_ablation_decoding} uses 64 prefill tokens and 256
generated tokens, with CPU-only as the baseline.

Buffer handoff adds much less time than language execution in
this workload. Importing a buffer takes about $1.2\,\mu$s across
the measured sizes, using allocation mapping rather than a payload
copy. A round trip for a $768$\,KB KV-sized block takes
$75.9\,\mu$s, compared with $109.0\,\mu$s for explicit copying.
At the $2.5$\,MiB payload used per request, the speedup reaches
$2.7\times$.

At the visual--language boundary, descriptor handoff for a $1.91$\,MB
visual-token batch takes $0.50\,\mu$s; serializing the payload takes
$142\times$ as long.

\begin{figure}[!t]
\centering
\subcapraggedrighttrue
\subfigure[Complete transport round trip]{%
\includegraphics{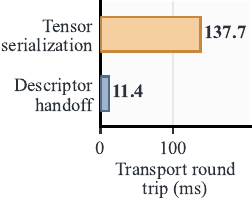}%
\label{fig:e2e_decoding}%
}\hfill
\subfigure[Average frame throughput]{%
\includegraphics{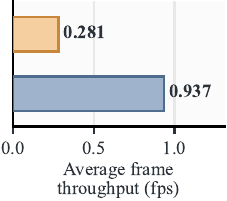}%
\label{fig:e2e_ttft}%
}
\caption{Visual-token transport on M4 Max: descriptor handoff versus
Python tensor serialization and reconstruction. (b) Frame throughput
under backlog.
}
\label{fig:e2e_ablation}
\end{figure}

Figure~\ref{fig:e2e_ablation}(a) measures the complete transport round
trip for an $800$\,KB visual payload. The Python byte path converts and
serializes the tensor, publishes and consumes the queue entry, and
reconstructs the tensor. Descriptor handoff passes the retained tensor's
reference and metadata. In panel (b), average frame throughput is input
frames divided by total run time; output throughput between the first
and last completed answers is $0.987$ and $0.983$\,fps, respectively.

\section{Runtime Failure Handling}
\label{app:failures}

An illegal slot transition in TABM, such as a second
publication or a release without a hold, fails the handoff instead of exposing stale
data. Lane faults are distinguished from request failures: a faulty
lane is removed from dispatch, whereas a failed request is retired
without affecting the lane that served it.

\section{QCS6490 Effective Read Bandwidth}
\label{app:qcs6490_bandwidth}

Figure~\ref{fig:engine_cost}(c) reports the 256\,MiB rates;
Table~\ref{tab:qcs6490_bandwidth} gives both input sizes, with each
engine reading complete 64 or 256\,MiB inputs alone or alongside the
others, as means and standard deviations over five runs.

\begin{table*}[t]
\caption{Effective read throughput on QCS6490~\cite{qualcomm_qcs6490_2025} (GB/s, mean $\pm$ SD over
five repeats) for engines reading alone and together.}
\label{tab:qcs6490_bandwidth}
\centering
\small
\setlength{\tabcolsep}{5pt}
\begin{tabular}{lrrrrrrrr}
\toprule
& \multicolumn{4}{c}{64\,MiB} & \multicolumn{4}{c}{256\,MiB} \\
\cmidrule(lr){2-5} \cmidrule(lr){6-9}
Configuration & CPU & GPU & NPU & Sum & CPU & GPU & NPU & Sum \\
\midrule
CPU & 15.20$\pm$0.06 & -- & -- & 15.20 & 15.52$\pm$0.04 & -- & -- & 15.52 \\
GPU & -- & 12.20$\pm$0.10 & -- & 12.20 & -- & 13.93$\pm$0.11 & -- & 13.93 \\
NPU & -- & -- & 10.61$\pm$0.10 & 10.61 & -- & -- & 11.68$\pm$0.06 & 11.68 \\
CPU+GPU & 4.63$\pm$0.09 & 11.07$\pm$0.09 & -- & 15.71 & 2.44$\pm$0.11 & 13.28$\pm$0.11 & -- & 15.72 \\
CPU+NPU & 8.63$\pm$0.10 & -- & 6.91$\pm$0.08 & 15.53 & 8.10$\pm$0.10 & -- & 7.81$\pm$0.09 & 15.90 \\
GPU+NPU & -- & 6.00$\pm$0.13 & 8.92$\pm$0.12 & 14.91 & -- & 5.91$\pm$0.04 & 9.73$\pm$0.02 & 15.63 \\
CPU+GPU+NPU & 5.92$\pm$0.10 & 4.89$\pm$0.06 & 4.83$\pm$0.04 & 15.65 & 5.17$\pm$0.02 & 5.32$\pm$0.00 & 5.32$\pm$0.00 & 15.80 \\
\bottomrule
\end{tabular}
\end{table*}

\section{Additional Context-Reuse and Streaming Results}
\label{app:context_results}

\subsection{Eight-Frame Stateful Execution}
\label{app:stateful_details}

The RK3588 panels in Figure~\ref{fig:stateful} use eight distinct frames from one OVO
video, arriving two seconds apart. Four different queries arrive at
$2.001$, $6.001$, $10.001$, and $14.001$\,s and bind
$C_2$, $C_4$, $C_6$, and $C_8$.
All six configurations execute fresh RKNN SigLIP~\cite{siglip} visual encoding and
the trained projector once per frame; each produces 197 visual rows.
The OneVision-0.5B W8 language model uses greedy decoding to generate
exactly 128 tokens per query, or 512 per session, in all six configurations.
Each configuration runs five times.

Configurations A--D fix OpenCL context preparation and CPU answer
generation. Without
reuse, each query rebuilds its 2/4/6/8-frame history; with reuse,
only new frames are appended. Both begin preparing frames as they
arrive. Disabling ingestion during generation delays subsequent
language preparation until the preceding answer finishes.

\begin{figure}[tb]
\centering
\includegraphics[width=\columnwidth]{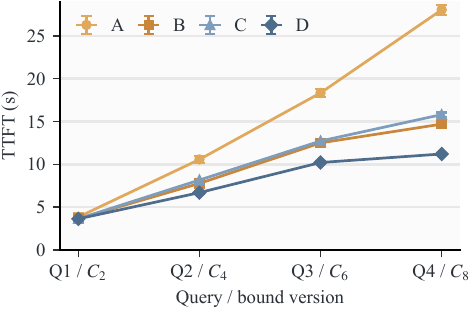}
\caption{Per-query TTFT for the four matched configurations in
Figure~\ref{fig:stateful}. A: neither mechanism; B: ingestion during
generation; C: context reuse; D: both (\name). Each query uses the same
bound version in all four configurations. Points and error bars show
five-run means and sample standard deviations.}
\label{fig:stateful_query_ttft}
\end{figure}

Figure~\ref{fig:stateful_query_ttft} shows how query waiting changes
as history grows. Mean TTFT falls from $15.20$ to $7.94$\,s with both
mechanisms. After reuse, ingestion during generation adds $15.6\%$
throughput. With both mechanisms, \name exceeds the reuse-enabled
CPU-only and GPU-only baselines by $11.6\%$ and $37.2\%$ in req/s,
with mean TTFT lower by $15.1\%$ and $34.6\%$, respectively.
Ingestion and answering overlap for $7.01\pm0.14$\,s.
All sixty paired comparisons preserve token IDs and first-step
logits as ingestion advances through the bound versions.

\begin{figure}[tb]
\centering
\includegraphics[width=\columnwidth]{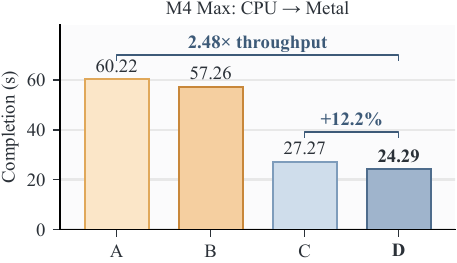}
\caption{CPU-to-Metal execution on M4 Max: $2.48\times$ request
throughput with context reuse and ingestion during generation
(D vs.\ A), and $12.2\%$ additional throughput after reuse (D vs.\ C).
Qwen2.5-VL-3B with 4-bit language weights processes eight frames and
four queries, generating 128 tokens per query. A--D use the mechanism
settings in Figure~\ref{fig:stateful}.}
\label{fig:mac_stateful}
\end{figure}

\noindent\textbf{M4 Max.}
Figure~\ref{fig:mac_stateful} uses the same eight frames and query
arrivals, with fresh MLX visual encoding and Qwen2.5-VL-3B Q4\_K\_M
language execution. CPU ingests the frames; Metal~\cite{apple_metal} imports each bound
context and generates the answer. Each frame supplies 196 visual
rows, and every configuration generates four 128-token answers.
The session time includes visual encoding, context construction,
state handoff, and generation.

\begin{table}[tb]
\centering
\caption{M4 Max CPU-to-Metal execution on the eight-frame stream.
A--D use the same mechanism settings as Figure~\ref{fig:mac_stateful}.}
\label{tab:mac_stateful}
\small
\begin{tabular}{lrr}
\toprule
Configuration & Completion (s) & Mean TTFT (s) \\
\midrule
A & 60.219 & 21.345 \\
B & 57.259 & 19.837 \\
C & 27.265 & 7.760 \\
D (\name) & \textbf{24.293} & \textbf{6.234} \\
\bottomrule
\end{tabular}
\end{table}

Reuse reduces frame prefills from twenty to eight. Ingestion during
generation adds $12.2\%$ throughput after reuse; together the two
mechanisms reduce mean TTFT by $70.8\%$ (Table~\ref{tab:mac_stateful}).
GPU-only execution with context reuse completes the same session in
$26.2$\,s, so D delivers $7.9\%$ more request throughput.
All twelve paired comparisons with the same path's sequential execution
match token sequences and complete first-continuation logit vectors.

\subsection{Cross-Query Multimodal Context Reuse}

Four requests share one image and alternate two questions, generating
eight tokens each. The first request builds the context; the remaining
three reuse it. Both platforms keep models resident, and the measured
window includes the first cache miss, image processing, visual
execution, language state preparation or import, and generation.
RK3588 uses five four-request runs per configuration.

On RK3588, RKNN SigLIP and the trained projector feed an OpenCL
OneVision-0.5B W8 context builder; CPU answers import its cached
24-layer KV state without replaying the language prefix. M4 Max uses
MLX Qwen2.5-VL vision and native Qwen2.5-VL-3B Q4\_K\_M Metal execution~\cite{llama_cpp,apple_metal},
retaining the root and forking a private suffix for each answer.
Visual encoding and common-prefix construction each fall from four
executions to one. All twenty RK3588 and four M4 Max cached answers
match their rebuild counterparts; RK3588's first-token logits also
match exactly.

\begin{figure*}[tb]
\centering
\includegraphics[width=\textwidth]{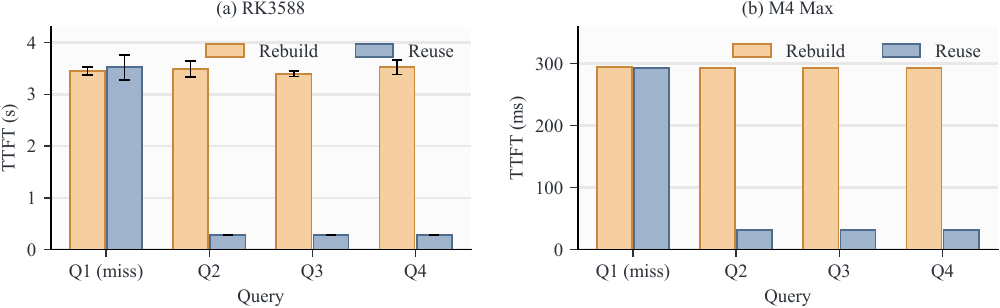}
\caption{Cross-query context reuse, including the initial cache miss.
Four requests share a visual prefix and generate eight tokens each.
Mean TTFT falls by $68.6\%$ on RK3588 and $67.0\%$ on M4 Max;
reuse delivers $2.79\times$ and $2.34\times$ the end-to-end request
throughput of rebuilding context, respectively. Subsequent requests
avoid visual encoding and common-prefix construction. RK3588 bars
show five-run means with sample-SD error bars.}
\label{fig:context_reuse_ttft}
\end{figure*}

Figure~\ref{fig:context_reuse_ttft} separates the initial miss from
subsequent reuse. RK3588's mean TTFT falls from $3465$ to $1089$\,ms,
and session time from $14.820$ to $5.316$\,s. On M4 Max, mean TTFT
falls from $293$ to $97$\,ms and session time from $1.372$ to
$0.587$\,s.

\noindent\textbf{Larger models.}
Qwen2.5-VL-3B and OneVision-7B use INT4 language weights on the
RK3588 CPU. Each answers four distinct questions sharing one image,
generating 32 tokens per question.

Reusing context raises Qwen2.5-VL-3B's request rate from $0.0350$ to
$0.1011$\,req/s and reduces mean first-token time from $25.29$ to
$6.74$\,s. Subsequent queries return their first token in
$0.52$--$0.58$\,s. For OneVision-7B, request rate rises from
$0.0479$ to $0.0908$\,req/s and mean TTFT falls from $13.73$ to
$4.20$\,s; subsequent queries reach the first token in
$0.90$--$1.07$\,s. All four Qwen2.5-VL-3B output sequences and all
four OneVision-7B first-step logit vectors match their rebuild
counterparts.

\subsection{Incremental Context Extension}

In three-query incremental sequences, context reuse reduces completion
time by factors of $1.6$--$2.4$ across the two platforms. On M4 Max,
TTFT for a query after the fourth frame
falls from $403.5$ to $162.2$\,ms. On OpenCL, all three answers from
the retained-context sequence match those from rebuilt context.

Native Metal context reuse also accelerates a three-frame, three-query
sequence by $1.62$--$1.63\times$ relative to rebuilding history for
each question. This Qwen2.5-VL-3B Q4\_K\_M measurement uses prepared
MLX visual embeddings and eight generated tokens per answer. All
three reused-context answers match rebuilding.

\subsection{Replay-Based Context Preparation}
\label{app:stateful_routing}

A separate three-frame RK3588 experiment prepares lane-local context
through replay and retained replicas, rather than state import.
With eager replicas and state-aware routing, the first two queries
run on CPU and the third on OpenCL. Relative to CPU-only execution,
Q3's arrival-to-first-token time falls from $7.447$ to $4.397$\,s
($41.0\%$), and the third frame's arrival-to-publication delay falls
from $7.269$ to $4.101$\,s. Request throughput rises by $4.67\%$,
from $0.220$ to $0.230$\,req/s. OpenCL prepares $C_2$ and $C_3$ while
CPU answers from $C_1$ and $C_2$, with $1.165$ and $0.539$\,s of native
prefill/decoding concurrency.

Each query binds its required version, including when that version is
requested after later ingestion. All sixteen same-backend sequential-reference
comparisons match every token ID and first-step logit: four configurations,
each with three timed answers and one historical-version answer.

\subsection{Concurrent Sequence Generation}
\label{app:concurrent_sweep}

\begin{figure}[tb]
\centering
\subfigure[Speedup over GPU-only]{
\includegraphics[width=0.455\columnwidth]{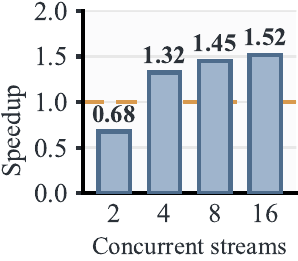}
}%
\subfigure[Median TTFT]{
\includegraphics[width=0.455\columnwidth]{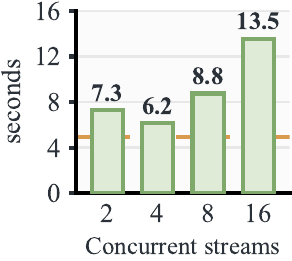}
}
\caption{RK3588 concurrent-stream sweep. Dashed lines mark the GPU-only
baseline shared by all four concurrency settings.}
\label{fig:rk_concurrent_sweep}
\end{figure}

The RK3588 sweep in Figure~\ref{fig:rk_concurrent_sweep} varies
concurrency from two to sixteen streams, reaching $52.0\%$ higher
request throughput than GPU-only execution. In the matched two-stream
comparison, each lane serves eight requests. The five-run
eight-stream comparison appears in Figure~\ref{fig:ovo_replay}, and \name
improves both throughput and p95 latency in every run.

On M4 Max, the OVO replay in Figure~\ref{fig:ovo_replay}(a,b) runs
Qwen2.5-VL-3B with Q4\_K\_M language weights. Answers are capped at 32
tokens, and each configuration generates 297 tokens in total.
With identical weights, precision, inputs, context, and decoding
settings, all 16 M4 Max OVO outputs match their sequential counterparts
in every configuration. All 80 RK3588 eight-stream comparisons match
the same lane running alone.

\subsection{SDEP Query Placement on RK3588 and QCS6490}
\label{app:context_placement}

Figure~\ref{fig:context_placement}(a,b) evaluates OneVision-0.5B W8 on
RK3588. RKNN SigLIP and the trained projector prepare
197 visual rows per frame before the query-placement window. The
CPU retains the requested context, and the GPU starts with a warm
model and reconstructs the query's context from retained embeddings.

Histories contain one, four, or eight frames. For each history,
short and medium backlogs use one background query $R_1$,
with $R_2$ arriving after 70\% or 30\% of its calibrated solo
duration. The long backlog queues three $R_1$ jobs and releases
$R_2$ after 10\% of the first job's solo duration. The corresponding
$R_1$ budgets are 50, 78, and 128 tokens for the three history
lengths; every $R_2$ generates 32 tokens.

Solo and co-running calibration supplies SDEP's preparation
and service estimates. The nine choices compare the two fixed policies
with SDEP. All 27 executions
match their own backend's sequential reference tokens and full first-step
logits.

\paragraph{Online dispatch on QCS6490.}
On QCS6490, two streams each contain
eight frames and four queries, with 128 output tokens per query.
OneVision-0.5B uses INT4 language weights; fresh visual encoding
runs through QNN~\cite{qnn_sdk} on the Hexagon NPU~\cite{qualcomm_hexagon}
with 8-bit weights and 16-bit activations, followed by the trained
CPU projector. All dispatch policies use the same arrival trace,
engine allocation, and initial cost estimates. Idle-first selects
an idle lane; reuse-first favors the lane holding the context;
SDEP selects the lane with the lowest $W+P+S$.

The dense trace releases frames at one-second intervals, offsets
the two streams by 50\,ms, and issues queries just after frames
2, 4, 6, and 8. Table~\ref{tab:qcs6490_dispatch} reports means over
five runs. Request throughput includes visual processing, context
preparation, state handoff, and generation; TTFT runs from query
arrival to the first token. \name reduces mean TTFT by $8.98\%$
relative to CPU-only with retained context. On sparse and burst
traces, the reductions are $17.91\%$ and $14.49\%$, respectively.

\begin{table}[t]
\centering
\caption{Online dispatch on QCS6490: request throughput and mean
TTFT over five runs of the dense two-stream trace. Both single-lane
baselines retain context.}
\label{tab:qcs6490_dispatch}
\small
\setlength{\tabcolsep}{6pt}
\begin{tabular}{lrr}
\toprule
Configuration & Throughput (req/s) & Mean TTFT (s) \\
\midrule
CPU-only & 0.15811 & 23.877 \\
GPU-only & 0.11250 & 31.741 \\
Idle-first & 0.14699 & 22.985 \\
Reuse-first & 0.10976 & 32.738 \\
\name & 0.15773 & \textbf{21.732} \\
\bottomrule
\end{tabular}
\end{table}

\subsection{Execution Modes with OneVision-7B and FastVLM-0.5B}
\label{app:multimodel_strategies}

Figure~\ref{fig:multimodel_strategies} uses INT4 language weights with
block size 64 and greedy generation on RK3588.
Model-specific visual embeddings are prepared before timing:
OneVision-7B uses its trained SigLIP/projector to produce 197 rows of
3584 values per frame, and FastVLM-0.5B~\cite{fastvlm2025} uses its
FastViTHD/projector to produce 256 rows of 896 values.
Table~\ref{tab:multimodel_strategies} gives the
absolute request rates and mean TTFT. The CPU and GPU lanes run at similar rates
with OneVision-7B, whereas FastVLM-0.5B runs faster on the CPU.

\begin{table}[t]
\centering
\caption{RK3588 execution modes with OneVision-7B and FastVLM-0.5B,
using precomputed visual embeddings. Each concurrent request generates
32 tokens; each stateful query generates 64. CPU+GPU sequential uses the
same ingestion, state-transfer, and answer paths as \name, but waits
for the previous answer before ingesting the next frame.}
\label{tab:multimodel_strategies}
\small
\setlength{\tabcolsep}{3pt}
\begin{tabular}{lrrrr}
\toprule
& \multicolumn{2}{c}{OneVision-7B} & \multicolumn{2}{c}{FastVLM-0.5B} \\
\cmidrule(lr){2-3} \cmidrule(lr){4-5}
Configuration & Req/s & \shortstack{Mean\\TTFT (s)}
& Req/s & \shortstack{Mean\\TTFT (s)} \\
\midrule
\multicolumn{5}{l}{\emph{Concurrent Sequence Generation}} \\
CPU-only   & 0.04664 & 46.592 & 0.53015 & 4.074 \\
GPU-only   & 0.04727 & 44.653 & 0.32787 & 6.565 \\
\name     & \textbf{0.06021} & \textbf{33.267} & \textbf{0.72757} & \textbf{3.073} \\
\midrule
\multicolumn{5}{l}{\emph{Stateful Streaming Generation}} \\
CPU-only       & 0.03517 & 27.622 & 0.39184 & 1.535 \\
GPU-only       & 0.03337 & 27.254 & 0.28519 & 2.079 \\
CPU+GPU sequential & 0.03427 & 26.930 & 0.38469 & 1.709 \\
\name         & \textbf{0.04304} & \textbf{21.471} & \textbf{0.43790} & \textbf{1.385} \\
\bottomrule
\end{tabular}
\end{table}

For Concurrent Sequence Generation, four independent visual requests
arrive together, each building its own context. The combined
configuration assigns two requests to each lane. Throughput uses the
shared batch wall clock, and TTFT starts at batch release, including
queueing. For Stateful Streaming Generation, two distinct frames
arrive at 0 and 2\,s, with a query 1\,ms after each frame, bound to $C_1$
and $C_2$ respectively. Single-lane configurations retain the history.
In the combined configuration, OpenCL ingests each frame and exports
its context for CPU import and generation. TTFT starts at each query's
scheduled arrival.

For OneVision-7B, the next frame's prefill overlaps the CPU's answer
to the previous query, and the second query's TTFT falls by $11.0$\,s,
from 40.83\,s with CPU-only to 29.84\,s.
All eight concurrent outputs match their own-lane references, and
all four stateful outputs match the sequential execution of the same
GPU-to-CPU path, including every first-step logit. Corresponding $C_1$
and $C_2$ prefixes have bit-identical per-layer K/V tensors under sequential
and overlapped execution: all 28 layers for OneVision-7B and all 24
for FastVLM-0.5B.

\subsection{Execution Modes on QCS6490}
\label{app:qcs6490}

On Qualcomm QCS6490~\cite{qualcomm_qcs6490_2025}, we evaluate
OneVision-0.5B, FastVLM-0.5B, and FastVLM-1.5B end to end, encoding
every frame on the device (Figure~\ref{fig:multimodel_strategies}).
The Qualcomm AI Engine Direct SDK (QNN)~\cite{qnn_sdk} runs the
visual encoders on the Hexagon NPU~\cite{qualcomm_hexagon} with 8-bit
weights and 16-bit activations, and the CPU runs the trained projectors.
OneVision-0.5B's SigLIP encoder executes as seven NPU partitions, and
pooling yields 197 rows per frame. FastVLM's full visual encoder runs
on the NPU, producing 256 rows per frame.
Language execution uses MNN~\cite{mnn} with INT4
weights on the CPU and, through OpenCL, on the Adreno
GPU~\cite{qualcomm_adreno}, with greedy generation. Every configuration
generates 128 tokens per query, with EOS termination disabled.

\begin{table}[t]
\centering
\caption{QCS6490 request throughput (req/s) with on-device visual
processing and 128 output tokens per query. CPU+GPU sequential is defined as in
Table~\ref{tab:multimodel_strategies}.}
\label{tab:qcs6490}
\small
\setlength{\tabcolsep}{4pt}
\begin{tabular}{lrrr}
\toprule
& OneVision & \multicolumn{2}{c}{FastVLM} \\
\cmidrule(lr){2-2} \cmidrule(lr){3-4}
Configuration & 0.5B & 0.5B & 1.5B \\
\midrule
\multicolumn{4}{l}{\emph{Concurrent Sequence Generation}} \\
CPU-only   & 0.2145 & 0.2143 & 0.0765 \\
GPU-only   & 0.1690 & 0.1622 & 0.0744 \\
\name     & \textbf{0.2453} & \textbf{0.2491} & \textbf{0.0897} \\
\midrule
\multicolumn{4}{l}{\emph{Stateful Streaming Generation}} \\
CPU-only       & 0.1545 & 0.1488 & 0.0572 \\
GPU-only       & 0.1112 & 0.0974 & 0.0522 \\
CPU+GPU sequential & 0.1418 & 0.1251 & 0.0540 \\
\name         & \textbf{0.1713} & \textbf{0.1553} & \textbf{0.0671} \\
\bottomrule
\end{tabular}
\end{table}

Concurrent Sequence Generation serves four requests with distinct
frames and questions. Stateful Streaming Generation ingests eight
frames arriving one second apart and answers four queries bound to $C_2$,
$C_4$, $C_6$, and $C_8$. As on RK3588, OpenCL ingests each frame and exports its
context, and the CPU imports it to answer.
Table~\ref{tab:qcs6490} gives the absolute request rates. The CPU is the
faster single lane for all three models. Ingestion
during generation raises request throughput by $20.8$--$24.3\%$ over
CPU+GPU sequential. Ingestion
overlaps decoding for $10.55$, $13.82$, and $26.51$\,s with OneVision-0.5B,
FastVLM-0.5B, and FastVLM-1.5B, respectively.

In all 24 comparisons, eight per model, outputs match token IDs and
complete first-step logits: each concurrent request matches its own
lane's reference, and each streaming answer matches sequential execution
on the same GPU-to-CPU path.

\begin{table}[t]
\centering
\caption{QCS6490 accuracy (\%) on 200 single-frame OVO questions per
model. CPU columns vary language-weight precision with the same visual
frontend. The last column uses INT4 language weights and GPU-built
context imported for CPU generation.}
\label{tab:qcs6490_accuracy}
\small
\setlength{\tabcolsep}{3pt}
\begin{tabular}{lrrrr}
\toprule
& \multicolumn{3}{c}{CPU} & GPU$\to$CPU \\
\cmidrule(lr){2-4}
Model & INT4 & INT8 & FP16 & INT4 \\
\midrule
OneVision-0.5B & 47.0 & 59.0 & 58.0 & 50.5 \\
FastVLM-0.5B   & 69.5 & 75.5 & 74.0 & 70.5 \\
FastVLM-1.5B   & 74.0 & 72.0 & 73.0 & 72.5 \\
\bottomrule
\end{tabular}
\end{table}

GPU-to-CPU generation achieves task accuracy comparable to CPU-local
INT4 (Table~\ref{tab:qcs6490_accuracy}), with KV imported for all
600 task queries.

\subsection{State Handoff}
\label{app:state_handoff}

\noindent\textbf{RK3588.}
Figure~\ref{fig:kv_import} compares CPU incremental replay with import
of ready OpenCL contexts for three OneVision-0.5B W8 queries. Each
frame has 197 prepared visual rows; each answer generates eight
tokens. Mean time to the first token falls from $1482$ to $319$\,ms. Repeated
queries of each imported version match every token and first-step
logit in all three comparisons.

\noindent\textbf{M4 Max.}
Table~\ref{tab:correctness}(a) uses Qwen2.5-VL-3B Q4\_K\_M and 200
single-frame OVO queries with fixed $384\times384$ images. MLX
prepares 196 visual tokens per image. CPU-local, Metal-local and
CPU-to-Metal paths receive the same visual embeddings, questions,
and option order. The CPU root is exported before the question;
Metal imports it, prefills the question, and emits one greedy choice
token.

\subsection{Integrated Multi-Stream Execution}
\label{app:integrated_multistream}

\begin{figure*}[t]
\centering
\includegraphics[width=\textwidth]{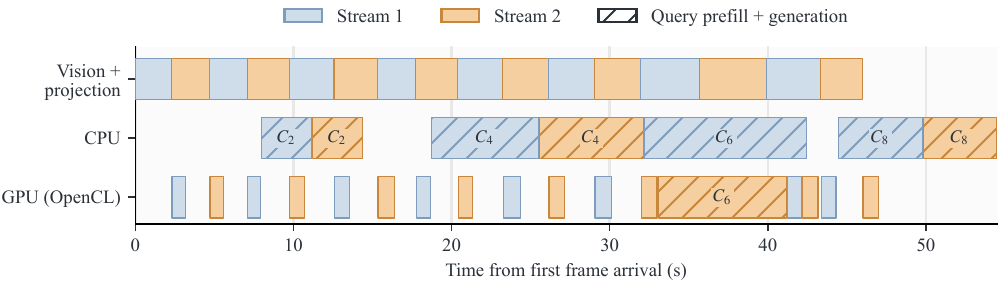}
\caption{Both execution modes in the EOS-terminated two-stream RK3588
configuration.
Colors identify the streams. Solid language intervals ingest frames;
hatched intervals process queries, including query prefill and
generation. Query labels identify the bound context version.
Vision and projection process all sixteen arriving frames.}
\label{fig:integrated_timeline}
\end{figure*}

The two-stream evaluation uses OneVision-0.5B with INT8 language weights and
FP16 SigLIP vision on RK3588. Two OVO video clips supply eight frames
each, and each query binds the context version covering the frames
that arrived before it. Generation is greedy. Each frame passes
through the NPU visual encoder and trained projector, then pooling
produces 197 visual rows.

All four configurations use the same arrival trace, and the timed
window includes visual encoding, projection, initial context
preparation, state handoff, queueing, and generation. \name
retains separate histories for the two streams and uses SDEP to schedule
context ingestion and query execution, with one shared queue per engine.
Published query-visible contexts remain fixed
as ingestion advances.

In the run of Figure~\ref{fig:context_placement}(c,d), every answer
generates 112 tokens and each configuration runs three times.
Each stream releases its eight frames at four-second intervals, with
the second stream starting 50\,ms after the first. Four queries per
stream arrive 1\,ms after frames 2, 4, 6, and 8 and bind $C_2$, $C_4$,
$C_6$, and $C_8$. Arrivals follow this schedule independently of
answer completion.
Table~\ref{tab:integrated_g112} gives request throughput and mean TTFT.
Against CPU reuse, \name lowers p95 TTFT from $27.64$ to $17.59$\,s,
p95 query completion from $31.62$ to $21.94$\,s, and session time from
$59.67$ to $49.89$\,s. Ingestion of new context overlaps decoding from
earlier versions for $5.80$\,s on average. Across the integrated runs,
all 168 answers reproduce the token IDs and complete first-step logits
of sequential execution along the same backend path.

\begin{table}[t]
\centering
\caption{Two-stream RK3588 request throughput (req/s) and mean TTFT
(s). Parentheses give throughput SD over three runs for the
fixed-112-token setting.}
\label{tab:integrated_g112}
\small
\setlength{\tabcolsep}{3.5pt}
\begin{tabular}{@{}lrrrr@{}}
\toprule
& \multicolumn{2}{c}{112 tokens} & \multicolumn{2}{c}{EOS, 128-token cap} \\
\cmidrule(lr){2-3} \cmidrule(lr){4-5}
Configuration & req/s & TTFT & req/s & TTFT \\
\midrule
Naive Parallelism & 0.1261\,{\scriptsize(0.0016)} & 17.01 & 0.118 & 16.96 \\
CPU, reuse & 0.1341\,{\scriptsize(0.0004)} & 17.27 & 0.134 & 15.11 \\
GPU, reuse & 0.1191\,{\scriptsize(0.0007)} & 19.37 & 0.119 & 18.30 \\
\name & \textbf{0.1604}\,{\scriptsize(0.0013)} & \textbf{11.25} & \textbf{0.147} & \textbf{12.14} \\
\bottomrule
\end{tabular}
\end{table}

The EOS-terminated configuration in Figure~\ref{fig:integrated_timeline}
uses the same arrival schedule and a 128-token cap.
\name raises request throughput by $9.3\%$ and cuts
mean TTFT by $19.6\%$ relative to CPU reuse, and gains $24.8\%$ over
Naive Parallelism. Seven answers import the GPU's KV state and run on CPU;
one runs on OpenCL. CPU and GPU decode different streams concurrently
for $7.66$\,s, and within a stream, new-frame ingestion also overlaps
an earlier answer. The longest TTFT falls from $26.89$\,s with CPU
reuse to $22.06$\,s. All eight answers in each configuration reproduce
the token IDs and complete first-step logits of sequential execution along
the same backend path.

\section{Stage Costs and Second-Lane Decoding on M4 Max}
\label{app:mac_lane}

These measurements run the Qwen2.5-VL-3B language model with Q4\_K\_M
weights on Apple M4 Max. Metal leads the CPU at every stage, by
$7.99\times$ in 1024-token prefill but only $1.69\times$ in decoding,
close to its $1.62\times$ lead in sequential reads
(Table~\ref{tab:mac_stage_costs}).

\begin{table}[t]
\centering
\caption{CPU and Metal on Apple M4 Max for the Qwen2.5-VL-3B language
model.}
\label{tab:mac_stage_costs}
\small
\setlength{\tabcolsep}{4pt}
\begin{tabular}{@{}lrrr@{}}
\toprule
Measurement & CPU & Metal & Metal/CPU \\
\midrule
Sequential read, 1\,GiB (GB/s) & 286.82 & 463.77 & $1.62\times$ \\
Prefill, 1024 tokens (tok/s) & 270.91 & 2163.41 & $7.99\times$ \\
Decode, 32-token prefix (tok/s) & 90.51 & 152.51 & $1.69\times$ \\
\bottomrule
\end{tabular}
\end{table}

Each lane decodes a separate sequence from a retained prefix, alone and
co-running, with three runs per configuration
(Table~\ref{tab:mac_corun}). At a 1024-token prefix, co-running lowers
Metal from $149.09$ to $140.53$\,tok/s while the CPU
contributes $59.11$\,tok/s, so the two lanes deliver $199.64$\,tok/s,
$33.9\%$ more than Metal alone. A stage-cost model calibrated before
this operating point was measured predicts $204.28$\,tok/s,
$2.32\%$ above the measured rate, and chooses
co-running. Figure~\ref{fig:lane_interference} marks the prediction. In
the prefill and generation sweep of Figure~\ref{fig:lane_gain_mac}, the
CPU lane serves $0.125$\,req/s against $1.768$\,req/s on Metal at 1024
prefill tokens and eight output tokens.

\begin{table}[t]
\centering
\caption{Decode throughput (tok/s) on Apple M4 Max with each lane
decoding from a retained prefix, alone and co-running. The gain is over
Metal alone.}
\label{tab:mac_corun}
\small
\setlength{\tabcolsep}{4pt}
\begin{tabular}{@{}lrrrr@{}}
\toprule
Prefix & CPU & Metal & CPU+Metal & Gain \\
\midrule
32 tokens & 90.51 & 152.51 & 208.16 & $36.5\%$ \\
1024 tokens & 82.89 & 149.09 & 199.64 & $33.9\%$ \\
\bottomrule
\end{tabular}
\end{table}

\section{State-Driven Execution Planning}
\label{app:sdep}

\begin{figure}[t]
\centering
\includegraphics[width=\columnwidth]{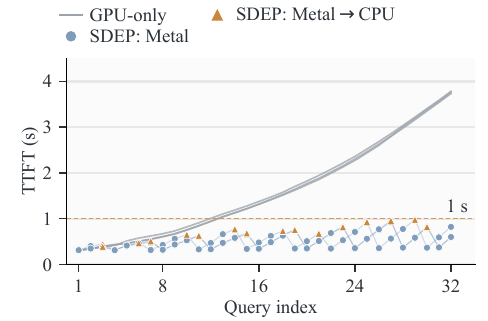}
\caption{M4 Max query TTFT and SDEP paths across three runs.
Metal$\to$CPU uses Metal prefill and CPU decoding.}
\label{fig:sdep_query_traces}
\end{figure}

The M4 Max live stream uses Qwen2.5-VL-3B with Q4\_K\_M language
weights, 32 frames arriving at 1\,fps, and 128 generated tokens per
query. One query arrives 1\,ms after each frame and binds the context
version ending at that frame, giving 32 queries per run. Frames and
queries arrive on this schedule independently of answer completion.
Core~ML visual encoding passes embeddings through IOSurface and
TABM into the language session. Metal retains the stream context and
prefills questions; phase-level state continuation lets the CPU import
an answer branch after query prefill and decode it.

The implementation updates the service deficit at each query arrival,
subtracting $90\%$ of the observed inter-query interval from predicted
ingestion and answer work. Offload credit counts source work avoided
minus export cost.

Against GPU-only execution with context reuse, \name lowers mean
TTFT by $67\%$, from $1.607$ to $0.528$\,s, raises answer rate by
$9.3\%$, and reduces p95 full-answer latency by $20.0\%$. All 96
queries receive their first token within one second.
Table~\ref{tab:sdep_live} reports mean and p95 TTFT. The \name results
average three runs. Adding phase-level state continuation to SDEP
reduces mean TTFT from $591$ to $528$\,ms and p95 full-answer latency
from $3.882$ to $3.643$\,s, while answer rate rises from $0.944$ to
$0.978$\,req/s.

\begin{table}[t]
\centering
\caption{Live-stream TTFT on M4 Max with context reuse.}
\label{tab:sdep_live}
\small
\begin{tabular}{@{}lrr@{}}
\toprule
Configuration & Mean (s)$\downarrow$ & p95 (s)$\downarrow$ \\
\midrule
GPU-only & 1.607 & 3.56 \\
SDEP without phase-level continuation & 0.591 & 1.21 \\
\name & \textbf{0.528} & \textbf{0.86} \\
\bottomrule
\end{tabular}
\end{table}

The backlog experiments use the same model, precision, and visual
pipeline, with all 32 frames and their queries available at time zero.
The three workloads use one query per frame with 64-token answers,
one query per four frames with 32-token answers, and one query per frame
with 128-token answers: 32, 8, and 32 queries, respectively.
Each query binds the history ending at its designated frame, and each
workload runs three times.

At the 64-token point, context handoff at the current version raises
throughput from $1.534$ to $1.92$\,req/s. Letting answers read their
bound versions while ingestion advances increases it to $2.03$\,req/s,
and completion-time dispatch reaches $2.078$\,req/s. Across the three backlog
loads, throughput gains reach $35\%$. The live-stream mechanism
sequence reduces p95 TTFT from $3.56$ to $2.34$\,s with completion-time dispatch,
to $1.21$\,s with rate-aware capacity, and to
$0.86$\,s with phase-level state continuation.

All 96 answers match their own sequential path in output tokens, root
positions, and first-continuation logits. All 96 visual-token handoffs
retain zero-copy execution through IOSurface and TABM.

\end{document}